%% file: main.tex
\documentclass{article} % For LaTeX2e
\usepackage{iclr2026_conference,times}

\input{math_commands.tex}

\usepackage{url}
\usepackage{xcolor}         % colors
\usepackage{amsmath}
\usepackage{amssymb}
\usepackage{caption}
\usepackage{graphicx}
\usepackage{dsfont}
\usepackage{wrapfig}
\usepackage{xspace}
\usepackage{titlecaps}
\usepackage[inline]{enumitem}
\usepackage{subcaption}
\usepackage{multirow}
\usepackage[ruled, noend]{algorithm2e}
\usepackage{booktabs}
\usepackage{wrapfig}
\usepackage{listings}
\usepackage{hyperref}
\usepackage{inconsolata}
\usepackage{booktabs,tabularx,array}
\newcolumntype{Y}{>{\raggedright\arraybackslash}X}

\newcommand{\dataset}{\ensuremath{\mathcal{D}}}
\newcommand{\parameters}{\ensuremath{\theta}}
\newcommand{\llm}{\ensuremath{\textbf{LLM}}}
\newcommand{\lbsfunc}{\ensuremath{\textbf{LBS}}}
\newcommand{\promptopt}{\ensuremath{\textbf{UPDATE}}}
\newcommand{\tglossfunc}{\ensuremath{\textbf{LOSS}}}
\newcommand{\tgbackward}{\ensuremath{\textbf{BACKWARD}}}
\newcommand{\tgstep}{\ensuremath{\text{optimizer}.\textbf{STEP}}}
\newcommand{\importancefunc}{\ensuremath{\textbf{F}}}
\newcommand{\xnew}{\ensuremath{x_\text{new}}}
\newcommand{\ynew}{\ensuremath{y_\text{new}}}
\newcommand{\posterior}{\ensuremath{p(\theta \mid \dataset)}}
\newcommand{\likelihood}{\ensuremath{p(\dataset \mid \theta)}}
\newcommand{\modeldens}{\ensuremath{p(y \mid x, \theta)}}
\newcommand{\prior}{\ensuremath{p(\theta)}}

\newcommand{\unnorm}{\ensuremath{g}}
\newcommand{\proposal}{\ensuremath{q}}
\newcommand{\thetaprop}{\ensuremath{\theta'}}
\newcommand{\priorstr}{\ensuremath{s}}

\newcommand{\appref}[1]{\hyperref[#1]{App.~\ref*{#1}}}

\newcommand{\mcmcalgfull}{Metropolis-Hastings through LLM Proposals\xspace}
\newcommand{\mcmcalg}{MHLP\xspace}
\newcommand{\lbs}{LLM-based system\xspace} % singular
\newcommand{\lbss}{LLM-based systems\xspace} % plural

\newlength{\itemizelength} 
\title{Textual Bayes: Quantifying Prompt \\ Uncertainty in LLM-Based Systems}

\author{%
  Brendan Leigh Ross\thanks{Equal contribution.} \quad No\"el Vouitsis$^*$ \quad Atiyeh Ashari Ghomi \\
  \textbf{Rasa Hosseinzadeh \quad Ji Xin \quad Zhaoyan Liu \quad Yi Sui \quad Shiyi Hou} \\
  \textbf{Kin Kwan Leung \quad Gabriel Loaiza-Ganem \quad Jesse C. Cresswell}\\
  \texttt{\{brendan, noel, atiyeh, rasa, zhaoyan,}\\
\texttt{amy, gloria, kk, gabriel, jesse\}@layer6.ai} \\
\texttt{ji.xin@uwaterloo.ca} \\
  Layer 6 AI, Toronto, Canada
}

\iclrfinalcopy % Uncomment for camera-ready version, but NOT for submission.
\begin{document}

\maketitle

\begin{abstract}
\input{sections/0-abstract.tex}
\end{abstract}

\section{Introduction}
\label{sec:introduction}
\input{sections/1-intro}

\section{Background and Terminology}
\label{sec:background}
\input{sections/2-background}

\section{Textual Bayes}
\label{sec:method}
\input{sections/3-method}

\section{Experiments}
\label{sec:results}
\input{sections/4-results}

\section{Related Work}
\label{sec:related}
\input{sections/5-related}

\section{Conclusions, Limitations, and Future Work}
\label{sec:conclusion}
\input{sections/6-conclusion}

% \subsubsection*{Author Contributions}
% If you'd like to, you may include  a section for author contributions as is done
% in many journals. This is optional and at the discretion of the authors.

% \subsubsection*{Acknowledgments}
% Use unnumbered third level headings for the acknowledgments. All
% acknowledgments, including those to funding agencies, go at the end of the paper.

\bibliography{bib}
\bibliographystyle{iclr2026_conference}

\clearpage
\appendix
\section{Method Details}
\label{app:method}
\input{sections/A-method.tex}

\section{Experiment Details}
\label{app:experiment}
\input{sections/B-experiment}

\end{document}

%% file: math_commands.tex
\usepackage{amsmath,amsfonts,bm}

\def\eqref#1{equation~\ref{#1}}
\def\1{\bm{1}}

\DeclareMathAlphabet{\mathsfit}{\encodingdefault}{\sfdefault}{m}{sl}
\SetMathAlphabet{\mathsfit}{bold}{\encodingdefault}{\sfdefault}{bx}{n}

\newcommand{\E}{\mathbb{E}}

%% file: sections/0-abstract.tex
Although large language models (LLMs) are becoming increasingly capable of solving challenging real-world tasks, accurately quantifying their uncertainty remains a critical open problem—one that limits their applicability in high-stakes domains. This challenge is further compounded by the closed-source, black-box nature of many state-of-the-art LLMs. Moreover, \lbss can be highly sensitive to the prompts that bind them together, which often require significant manual tuning (i.e., prompt engineering). In this work, we address these challenges by viewing \lbss through a Bayesian lens. We interpret prompts as textual parameters in a statistical model, allowing us to use a small training dataset to perform Bayesian inference over these prompts. This novel perspective enables principled uncertainty quantification over both the model’s textual parameters and its downstream predictions, while also incorporating prior beliefs about these parameters expressed in free-form text. To perform Bayesian inference—a difficult problem even for well-studied data modalities—we introduce \mcmcalgfull (\mcmcalg), a novel Markov chain Monte Carlo (MCMC) algorithm that combines prompt optimization techniques with standard MCMC methods. \mcmcalg is a turnkey modification to existing LLM pipelines, including those that rely exclusively on closed-source models. Empirically, we demonstrate that our method yields improvements in both predictive accuracy and uncertainty quantification (UQ) on a range of LLM benchmarks and UQ tasks. More broadly, our work demonstrates a viable path for incorporating methods from the rich Bayesian literature into the era of LLMs, paving the way for more reliable and calibrated LLM-based systems.

%% file: sections/1-intro.tex
Large language models (LLMs) have become increasingly embedded in our daily lives, with growing adoption across domains such as customer support \citep{chaturvedi2023opportunities}, code generation \citep{wang2021codet5,chen2021evaluating}, scientific research \citep{boiko2023emergent,schmidgall2025agent,yamada2025ai}, and creative writing \citep{gomez2023confederacy}. As their capabilities continue to advance, there is also mounting interest in deploying them in agentic systems, wherein they perform tasks autonomously on behalf of users \citep{wooldridge1995intelligent,xi2025rise}. Despite their proliferation, however, trust in LLMs remains limited, largely due to their propensity to generate hallucinated content \citep{maynez2020faithfulness, xu2024hallucination} and their susceptibility to adversarial attacks and jailbreaking \citep{wei2023jailbroken,zou2023universal,yan2024backdooring}. These vulnerabilities in \lbss therefore must be addressed, especially to fully unlock high-stakes domains such as finance and medicine. A key step towards mitigating these risks is to reliably quantify the uncertainty of \lbss. Accurate measures of uncertainty ensure that, when unable to answer, \lbss can abstain, defer to human experts, or augment their context with subroutines based on retrieval or reasoning \citep{lewis2020retrieval,wei2022chain}.

\begin{figure}[t]
   \centering
   \makebox[\textwidth][c]{%
     \includegraphics[width=0.95\textwidth]{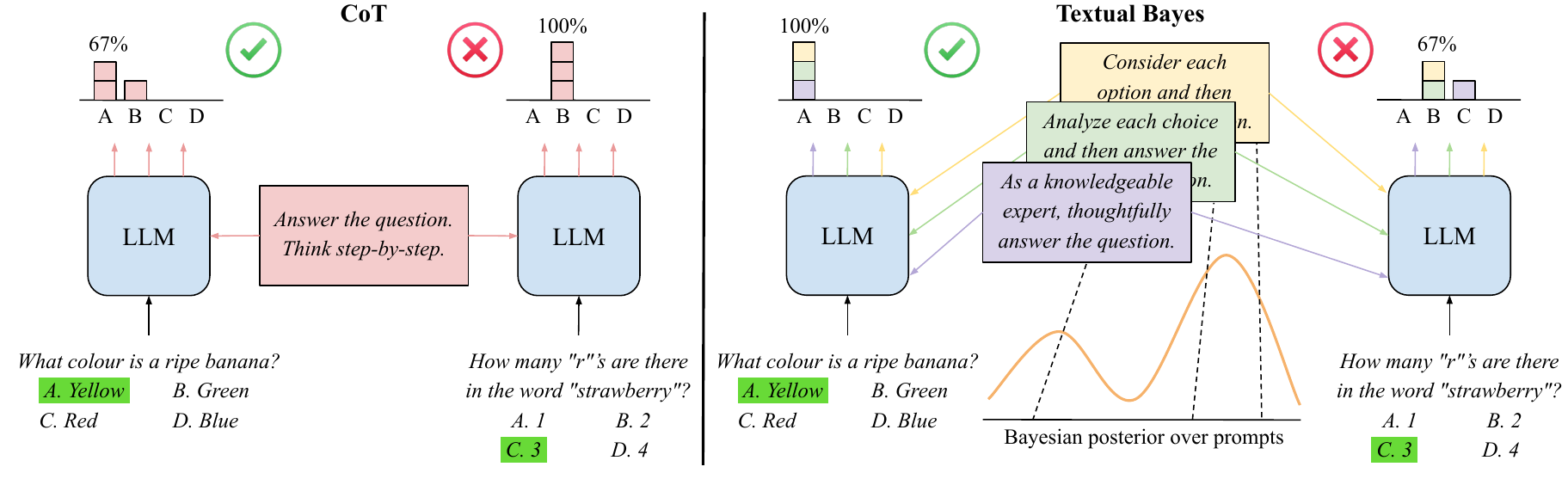}
   }
   \caption{\footnotesize{In chain-of-thought (CoT) prompting (left), answers are generated by an LLM using a single fixed prompt; this frequentist approach does not account for uncertainty about how the model should be prompted, causing potential issues such as overconfidence on incorrect answers. In Textual Bayes (right), we sample prompts from our Bayesian posterior and use each to generate answers from the LLM; this allows for principled uncertainty quantification over both the prompts themselves and the resulting generated answers.}}
   \label{fig:main_fig}
   %\vspace{-20pt}
   \vspace{-10pt}
\end{figure}

Despite recent progress, uncertainty quantification (UQ) for LLMs is far from solved and no consensus exists over exactly what should be quantified \citep{kuhn2023semantic, wang2024subjective, yang2024verbalized}. In this work, we propose to better quantify uncertainty in \lbss by viewing them through a Bayesian lens. In light of a model, observational data, and one's prior beliefs, Bayesian inference uses Bayes' rule to compute the distribution of possible model parameters.
Commonly applied in classical statistics and deep learning, Bayesian inference is a principled and mathematically grounded approach to UQ \citep{bernardo2009bayesian}.  Bayesian techniques have led to high-profile successes in methods like variational autoencoders \citep{kingma2014auto} and Bayesian neural networks \citep{blundell2015weight}. As with their 20th century forebears (e.g., \citep{duane1987hybrid,saul1996mean}), these methods estimate uncertainty over high-dimensional continuous variables. Here, we bring Bayesian methods into the age of LLMs. In \lbss, the main variables of interest are prompts, since LLMs themselves are often black boxes that can only be accessed via an API. By treating prompts as textual parameters in a statistical model, as illustrated in \autoref{fig:main_fig}, we can use Bayesian inference to estimate distributions over their values. These distributions rigorously quantify our uncertainty about the models themselves. Furthermore, they can be integrated into uncertainty estimates on the system's downstream outputs via easy-to-compute Monte Carlo estimates. To the best of our knowledge, we are the first to perform Bayesian inference over the space of free-form prompts in \lbss.

Adapting Bayesian methods to text has its challenges and advantages. On the one hand, textual variables are discrete, making it difficult to apply traditional Bayesian deep learning techniques such as gradient-based Markov chain Monte Carlo (MCMC) \citep{welling2011bayesian} or variational inference \citep{saul1996mean}. We address this obstacle with a novel text-based MCMC method: \emph{\mcmcalgfull} (\mcmcalg). On the other hand, textual variables are better suited conceptually to Bayesian modelling than high-dimensional continuous variables such as the weights of a deep neural network. Bayesian inference famously requires the specification of prior beliefs about a variable; textual variables are more amenable to human priors than neural network weights, and as we show, prior beliefs can be readily incorporated into \lbss as free-form text.

To advance and justify our proposed method, this work contains the following contributions:\vspace{-5pt}
\begin{enumerate}
\setlength{\leftskip}{\itemizelength}
    \item We take a novel perspective on \lbss in which prompts are viewed as Bayesian textual parameters $\parameters$ in a model $p(y \mid x, \parameters)$. We show how this formulation leads to a principled way to incorporate prior beliefs about $\theta$ while quantifying our inherent uncertainty in the model.
    \item To implement our Bayesian approach, we propose \emph{\mcmcalgfull} (\mcmcalg), an MCMC algorithm to sample from intractable distributions over textual variables. \mcmcalg has broad potential applications even beyond Bayesian inference.
    \item We propose a novel metric of model calibration, \emph{semantic expected calibration error}, for quantifying calibration, a form of UQ, on free-form textual outputs.
    \item We systematically evaluate our method through standard LLM benchmarks and baselines, showing that it improves performance while providing state-of-the-art UQ over model outputs.
\end{enumerate}

%% file: sections/2-background.tex
\subsection{LLM-Based Systems}\label{sec:lbs}
The central object in this work is the \lbs. The most common \lbs is one consisting of a single input $x$ (e.g., a question), a prompt $\theta$ (e.g., a system message defining instructions for the model's behaviour, shared across all $x$), and an output $y$ (e.g., the model's predicted answer), which we denote
\begin{equation}\label{eq:single_llm_system}
    y = \llm(x; \theta).
\end{equation}
We allow $\llm(x; \theta)$ to be any open- or closed-source model that we view as a \emph{random} function of $x$ and $\theta$, whose randomness depends on the underlying LLM sampling strategy (e.g., greedy, temperature, nucleus). In general throughout the work, capitalized, boldface function names will indicate random functions comprising one or more LLM calls.

\lbss can be more complex than single-prompt models. Many recent works have proposed to group LLM calls of arbitrary count and complexity into pipelines parameterized by the prompts used at each step \citep{khattab2024dspy,zhugegptswarm,yuksekgonul2025optimizing,cheng2024trace,hu2024automated}. For example, Self-Refine \citep{madaan2023self} iterates on an initial LLM output by alternating between an LLM call providing feedback and one incorporating the feedback into refinement. Fully agentic systems integrate multiple LLM and tool calls to arrive at a final output. In full generality, we can describe a forward pass through an \lbs as 
\begin{equation}
    y = \lbsfunc(x ; \theta),
\end{equation}
where $\lbsfunc(\cdot ; \theta)$ can be described as a directed acyclic graph with $k$ edges in which each edge $e_j$ corresponds to an LLM call $\llm({ \cdot }; \theta_j)$ parameterized by a prompt $\theta_j$ and where we denote the combination of all prompts in the system as $\theta= (\theta_1, \ldots, \theta_k)$. Since each LLM call in the system is potentially random, $y$ is a random function of $x$ parameterized by $\theta = (\theta_1, \ldots, \theta_k)$. The \lbs thus forms a statistical model for $y$ whose probability mass function we express as $\modeldens$, where sampling $y \sim \modeldens$ is equivalent to computing $y = \lbsfunc(x ; \theta)$.

Unlike a linear regressor or neural network where $\theta$ denotes continuous model parameters, for an \lbs $\theta$ denotes \emph{textual} parameters.
From the statistical modelling perspective, a natural next step is to find the optimal value of $\theta$. For example, given an i.i.d.\ dataset $\dataset = \{(x_1, y_1), \ldots, (x_n, y_n)\}$, one might want to perform maximum-likelihood:
\begin{equation}
\label{eq:mle}
    \theta^* 
    = \arg\max_\theta p(\dataset \mid \theta)
    = \arg\max_\theta \prod_{i=1}^n p(y_i \mid x_i, \theta).
\end{equation}
The discrete nature of textual parameters prevents us from applying gradient-based algorithms to maximize likelihood in \lbss. 
Prompt engineering can be understood as approximating $\theta^*$ by having a human propose candidates $\theta$ until adequate performance on a small dataset is reached. However, this manual process lacks rigour, is lengthy and tedious, and does not scale well.  

Past works have proposed heuristic approaches to automatically optimize prompts $\theta$ in \lbss \citep{zhou2022large,khattab2024dspy,zhugegptswarm,cheng2024trace}. Here, we focus on iterative prompt optimization methods, which we can express mathematically as a stochastic update function $\promptopt$ applied iteratively to an initial prompt $\theta^{(0)}$:
\begin{equation}\label{eq:update}
    \theta^{(t)} = \promptopt(\theta^{(t-1)}).
\end{equation}
For simplicity, we assume that $\promptopt$ is Markovian; i.e., not a function of $\theta$ values from earlier than $t-1$. $\promptopt$, which consists of one or more LLM calls, is itself an \lbs.

One particularly relevant prompt optimization method is TextGrad \citep{yuksekgonul2025optimizing}. The TextGrad framework conceptualizes constructive feedback on prompts as \emph{textual gradients} and proposes a method for ``backpropagating'' feedback through an \lbs akin to backpropagation in neural networks. Although this framework is highly analogous to backpropagation of gradients for continuous variables, it does not formally optimize model likelihood.

\subsection{Bayesian Inference}
In this section, we briefly review Bayesian inference. For a more in-depth introduction, we refer the reader to \citet{mackay2003information}. Here, we allow $\modeldens$ to be any statistical model for some variable $y$ given another variable $x$. From the Bayesian perspective, there is uncertainty about the true value of $\theta$, and hence the point estimate $\theta^*$ given by maximum-likelihood may be an overly reductive way of summarizing a dataset $\dataset$. Bayesian statistics provides a formal way of capturing this uncertainty.

First, we encode our prior uncertainty (beliefs) about the true value of $\theta$ as a \emph{prior distribution} $p(\theta)$. Then, having observed a dataset $\dataset$, we update our beliefs about $\theta$ using Bayes' rule as
\begin{equation}\label{eq:bayes}
    p(\theta \mid \dataset) 
    = \frac{p(\theta)p(\dataset \mid \theta)}{p(\dataset)}
    % = \frac{p(\theta)\prod_ip(y_i \mid x_i, \theta)p(x_i)}{\int_{\theta'} p(\theta')\prod_i p(y_i \mid x_i, \theta)'p(x_i) d\theta'}
    = \frac{p(\theta)\prod_ip(y_i \mid x_i, \theta)}{\sum_{\theta'} p(\theta')\prod_i p(y_i \mid x_i, \theta')}.
\end{equation}
The \emph{posterior distribution} $p(\theta \mid \dataset)$ formally captures our uncertainty about $\theta$ in light of $(i)$ our prior beliefs and $(ii)$ the observed data. Given $p(\theta \mid \dataset)$ and a new unobserved datapoint $\xnew$, we can compute the \emph{posterior predictive distribution} of $\ynew$
via 
\begin{equation}\label{eq:post_pred}
    p(\ynew \mid \xnew, \dataset) 
    = \sum_\theta p (\ynew \mid \xnew, \theta) p(\theta \mid  \dataset)
    = \E_{\theta \sim p(\theta \mid \dataset)}\left[p (\ynew \mid \xnew, \theta)\right].
\end{equation}
\autoref{eq:post_pred} formalizes predictive uncertainty in terms of uncertainty over $\theta$. Since it is expressed as an expectation, we can estimate it via Monte Carlo sampling with draws from $p(\theta \mid \dataset)$.
The posterior predictive has immediate practical value: its value represents confidence in the prediction $\ynew$, and its variability (as measured by, e.g., variance or entropy) formally quantifies uncertainty.

The central challenge of Bayesian inference thus lies in sampling from the posterior $p(\theta\mid \dataset)$. As $\modeldens$ or $p(\theta)$ acquire even moderate complexity, sampling from $p(\theta\mid \dataset)$ quickly becomes intractable. In deep learning, Bayesian inference requires approximations such as gradient-based MCMC \citep{welling2011bayesian}, variational inference \citep{blundell2015weight}, or Laplace approximations \citep{ritter2018}. We highlight that all of these approaches rely on the differentiability of $\prior \likelihood$ with respect to $\theta$, so none can be readily applied to the context where $\theta$ is a discrete prompt in an \lbs.

\subsection{Markov Chain Monte Carlo and the Metropolis-Hastings Algorithm}

In Bayesian statistics, MCMC algorithms are a common technique for tractably sampling from the posterior $\posterior$ when only its numerator in \autoref{eq:bayes} can be computed for any values of $\theta$ and $\dataset$. First, fix $\dataset$ and let $\unnorm(\theta) = \prior\likelihood$ be the numerator of \autoref{eq:bayes}. Given an unnormalized density like $\unnorm(\theta)$, an MCMC algorithm is a general-purpose technique that specifies a Markov chain $\theta^{(1)}, \theta^{(2)}, \ldots, \theta^{(t)}, \ldots$ whose distribution converges to $\frac{\unnorm(\theta)}{\sum_{\theta'}\unnorm(\theta')}= p(\theta \mid \dataset)$ as $t \to \infty$. In practice, by generating enough samples from the Markov chain, we can approximate sampling from $\posterior$ without needing to evaluate it.

\begin{wrapfigure}[16]{r}{0.47\textwidth}
\vspace{-20pt}
\begin{minipage}{0.47\textwidth}
\begin{algorithm}[H]
\caption{Metropolis-Hastings}\label{alg:metropolis_hastings}
\textbf{Require:} $\theta^{(0)}$, $\proposal(\thetaprop \mid \theta)$, $g(\theta)$\;
\For{$t \gets 1$ \KwTo $T$}{
    Sample proposal: $\thetaprop \sim \proposal(\thetaprop \mid \theta^{(t-1)})$\;
    Compute acceptance probability:
    \[
    \gamma = \min\left(1, \frac{\unnorm(\thetaprop)\, \proposal(\theta^{(t-1)}\mid\thetaprop)}{\unnorm(\theta^{(t-1)})\, \proposal(\thetaprop\mid\theta^{(t-1)})}\right);
    \]
    Sample random number: $u \sim \text{Uniform}(0,1)$\;
    \eIf{$u < \gamma$}{
        Accept: $\theta^{(t)} \gets \thetaprop$\;
    }{
        Reject: $\theta^{(t)} \gets \theta^{(t-1)}$\;
    }
}
\KwRet{$\{\theta^{(t)}\}_{t=1}^T$}\;
\end{algorithm}
\end{minipage}
\vspace{-2em}
\end{wrapfigure}

The \emph{Metropolis-Hastings} algorithm (MH) is a generic and broadly applicable form of MCMC (for an introduction, see \citet{robert2015metropolis}). Starting with an initial sample $\theta^{(0)}$, MH iterates from sample $\theta^{(t-1)}$ to $\theta^{(t)}$ by generating a new \emph{proposal} $\thetaprop$ from a pre-defined \emph{proposal distribution} $\proposal(\thetaprop \mid \theta)$ and then either accepting it (i.e., setting $\theta^{(t)} := \thetaprop$) or rejecting it (i.e., setting $\theta^{(t)} := \theta^{(t-1)}$) based on an acceptance probability $\gamma$ (\autoref{alg:metropolis_hastings}).

In MH, the main ``tuneable hyperparameter'', the choice of proposal distribution $\proposal(\thetaprop \mid \theta)$, is constrained only by very mild regularity conditions. However, the choice of $\proposal$ has a pronounced effect on the practicality of the algorithm, with poor choices (e.g., ones that perturb $\theta$ too mildly or too strongly at each step) taking an intractable amount of time to converge to the limiting distribution $\posterior$. The importance of $\proposal$ is such that some of the most popular MCMC algorithms (e.g.,  Langevin Monte Carlo and Hamiltonian Monte Carlo \citep{duane1987hybrid,neal1996bayesian}) are simply special cases of MH with highly specialized choices of $\proposal$. Our method, \mcmcalg, will also fall into this category, being specialized for textual parameters $\theta$. The choice of $\proposal$ should be driven by any information available about the desired limiting distribution $\posterior$ \citep{rosenthal2011optimal}. Indeed, the optimal $\proposal(\thetaprop \mid \theta)$ would be \textit{equal} to the desired limiting distribution itself; if this were possible, of course, there would be no need to run MH in the first place. Nevertheless, we apply this intuition in \autoref{sec:method} as we adapt MH to textual data.

%% file: sections/3-method.tex
In this section, we describe our method for Bayesian inference on \lbss. We begin with the setup described in \autoref{sec:lbs}: an \lbs $\lbsfunc(x; \theta)$ that gives rise to a statistical model $p(y \mid x, \theta)$, where $x$ is the input, $y$ is the output, and $\theta = (\theta_1, \ldots, \theta_k)$ represents all the textual parameters involved in the system. We assume that $\modeldens$ can be evaluated for any  \lbs we consider; to do this in full generality for closed source models, we require some approximations, including selective use of open-source likelihoods as surrogates --- see \appref{app:approximations} for more details. 
Our Bayesian inference algorithm will provide samples $\theta^{(1)}, \ldots, \theta^{(m)} \sim\posterior$, which can in turn be used to quantify uncertainty over the system's outputs as per \autoref{eq:post_pred}.

\textbf{Textual priors}\quad To perform Bayesian inference, we must specify our prior beliefs about $\theta$ in the form of a distribution $p(\theta)$. Although $\theta$ lies in an infinite and semantically complex space of discrete text, humans are well equipped to reason and express their beliefs about textual variables. For example, a practitioner's prior about a prompt $\theta_j$ might be that it should describe the purpose of the corresponding LLM call, guidelines for how to solve the task at hand, and the expected structure of the output. To exploit this knowledge, we codify our beliefs about each parameter $\theta_j$ as a free-form human-written string of textual constraints $\priorstr_j$, and provide it to an LLM to model the resulting parameter as
\begin{equation}
\theta_j = \llm(s_j; \texttt{"Generate an LLM prompt satisfying the given constraints."}).
\end{equation}
% The probability assigned by this call to a particular $\theta_i$ represents the probability $\theta_i$ satisfies the textual constraints encoded by $s_i$.
For simplicity, we construct our prior $p(\theta) = \prod_{j=1}^kp(\theta_j)$ by assuming that all textual variables are independent, but this setup can be easily generalized by specifying joint constraints over multiple parameters $\theta_j$ and modelling them in a single LLM call.

\textbf{\mcmcalgfull}\quad Having constructed our prior $p(\theta)$, we now need an algorithm to sample from $p(\theta \mid \dataset)$. A generally applicable MCMC method for text could have wide-ranging applications even beyond Bayesian inference. To this end, we propose \mcmcalgfull (\mcmcalg), a text-specific variant of MH.

At the heart of \mcmcalg is our proposal distribution. We could in theory achieve the correct limiting distribution through almost any arbitrary choice of $\proposal(\thetaprop \mid \theta)$, like randomly replacing letters or words in $\theta$. But it is easy to see that such a proposal would rarely change $\theta$ semantically and never converge in practice. Instead, to generate useful proposals, we turn to LLMs. Analogously to how Langevin Monte Carlo uses gradient computation to exploit differentiable structure on $\posterior$, \mcmcalg uses LLM calls to exploit linguistic structure on $\posterior$.
Ideally,  $\proposal(\thetaprop \mid \theta)$ should be as similar to $p(\theta \mid \dataset)$ as possible. By this standard, as per the relationship $p(\theta \mid \dataset) \propto p(\dataset \mid \theta) p (\theta)$, samples $\thetaprop \sim \proposal(\thetaprop \mid \theta)$ should roughly satisfy the following criteria: $(i)$ $\thetaprop$ should satisfy all the constraints embodied by the prior $p(\thetaprop)$, and $(ii)$ $\thetaprop$ should provide strong downstream performance on $\dataset$ as measured by $p(\dataset \mid \thetaprop)$.

We take inspiration from the prompt optimization methods discussed in \autoref{sec:lbs} 
and use suggestions from LLMs to propose values of $\thetaprop$ that implement these guidelines.
The observation underpinning \mcmcalg is that iterative prompt optimization methods can be used to propose high-quality candidates $\thetaprop$. Here we recall our formalization of prompt optimization as an iterated stochastic update function $\promptopt$ (\autoref{eq:update}), and sample from $\proposal(\thetaprop \mid \theta)$ by computing $\thetaprop = \promptopt(\theta)$.\footnote{Some prompt optimization methods, such as the momentum variant of TextGrad \citep{yuksekgonul2025optimizing}, make updates based on a history of multiple past $\theta$ values. \mcmcalg can take advantage of such methods by running multiple steps of the optimizer per accept/reject decision, akin to Hamiltonian Monte Carlo.} 
Note that since $\promptopt$ is itself an \lbs just like our model $\lbsfunc$, and so, like the model density $\modeldens$, the value of $\proposal(\thetaprop \mid \theta)$ can be estimated by using an open-source model for the final LLM call of $\promptopt$ and by using the approximations in \appref{app:approximations}. 
Although \mcmcalg is agnostic to the underlying prompt optimization method, we use TextGrad \citep{yuksekgonul2025optimizing} in our implementation. By analogy to numerical losses in standard gradient-based optimization, TextGrad optimizes objectives described in natural language. We can thus express criteria $(i)$ and $(ii)$ as objectives in natural language and use TextGrad to propose improvements to $\theta$ based on these criteria. This choice of objectives is specific to Bayesian inference, but in \mcmcalg they can be easily replaced or modified to suit any textual distribution, which broadens its potential impact. We demonstrate one such example in \autoref{sec:conformal}.

\textbf{Method summary} We define MHLP as the variant of MH (\autoref{alg:metropolis_hastings}) acting on textual parameters $\theta$ in which the proposal step $\thetaprop \sim \proposal(\thetaprop \mid \theta^{(t-1)})$ is defined as a prompt optimization update $\thetaprop = \promptopt(\theta^{(t-1)})$. For our experiments, we implement $\promptopt$ as a TextGrad step (\autoref{alg:textgrad_update}). Additionally, as is common practice in Bayesian deep learning (e.g., \citet{blundell2015weight,daxberger2021laplace}), we employ approximations for tractability, including a tempered posterior \citep{wenzel2020good} and stochastic minibatch estimates of certain quantities in \autoref{alg:metropolis_hastings} (rather than evaluating them exactly). Due to space constraints, approximation details are relegated to \appref{app:approximations}.

{Having a collection $\{\theta^{(r)}\}_{r=1}^m \sim p(\theta \mid \dataset)$ of prompt samples, we can now put them to use at inference time to sample from the predictive posterior (\autoref{eq:post_pred}). Given an input $\xnew$, we can generate a set of samples $\{\ynew^{(r)}\}_{r=1}^m \sim p(\ynew \mid \xnew)$ via
\begin{equation}
    \theta^{(r)} \sim p(\theta \mid \dataset),
    \hspace{20pt}
    \ynew^{(r)} \sim p(\ynew \mid \xnew, \theta^{(r)});
\end{equation}
that is, by running $\ynew^{(r)} = \lbsfunc(\xnew; \theta^{(r)})$ for each sampled prompt $\theta^{(r)}$. The variability of the resulting answer set $\{\ynew^{(r)}\}_{r=1}^m$ can be interpreted as the uncertainty of the \lbs.

%% file: sections/4-results.tex
In this section, we empirically evaluate our proposed Textual Bayes method. Specifically, we aim to answer the following question: how does Bayesian inference on the prompts of an \lbs with our \mcmcalg algorithm translate into the system's downstream \textit{predictive performance} and UQ abilities? In \autoref{sec:qa_eval}, we demonstrate that our method outperforms comparable baselines in accuracy, calibration, and abstention capabilities on challenging LLM benchmarks. In \autoref{sec:conformal}, we adapt Textual Bayes to reducing hallucinations with conformal factuality \citep{mohri2024factuality}, a distinct context from traditional Bayesian inference. %using the conformal factuality framework \citep{mohri2024factuality}.

\textbf{Implementation}\quad For each dataset, we use \mcmcalg to generate samples $\theta^{(1)}, \ldots, \theta^{(m)} \sim p(\theta \mid \dataset)$ from a Markov chain of length $T$. To increase sample diversity we employ burn-in, in which a fixed number $d$ of initial MCMC samples are discarded, and thinning, in which we take every $h$-th sample thereafter until $m$ samples are obtained. Given a datapoint $\xnew$, we sample values of $\ynew$ using \autoref{eq:post_pred} and quantify uncertainty on the basis of these downstream outputs. Because our research focus is on compatibility with black-box LLMs, in this section we present experiments with GPT-4o or, when possible, GPT-4o-mini, depending on the difficulty of the dataset. For
details such as settings of $d$, $h$, $m$, and other dataset-specific hyperparameters, see \appref{app:experiment}.

\subsection{Uncertainty quantification with Textual Bayes}
\label{sec:qa_eval}

\textbf{Setup}\quad 
We consider the canonical \lbs consisting of a single LLM as defined by \autoref{eq:single_llm_system}.
Hallucinations in such systems occur when a model responds confidently with incorrect or ungrounded information, an issue that can be combatted with calibration \citep{kadavath2022language, wei2024measuring}. Calibration refers to the quality of a model's confidence score, or the probability it assigns to the correctness of its provided answer; in other words, how well the model ``knows what it knows''. Here, we test calibration in downstream responses resulting from Bayesian inference over the LLM's prompt. We compute confidences by generating 10 responses from each system and measuring the frequency of each response. For \mcmcalg, we initialize $\theta^{(0)}$ to be a generic chain-of-thought (CoT) \citep{wei2022chain} prompt: \texttt{"Answer the question. Think step-by-step."}.

\textbf{Baselines}\quad We compare our method against four frequentist baselines. \emph{Paraphrasing} and \emph{System-Message} are two prompt perturbation methods proposed by \citet{gao2024spuq}. These methods inject prompt stochasticity by rephrasing the question or system prompt in a question-answering context.\footnote{Our implementation has minor differences from the cited paper. For further details see~\appref{app:baseline}.} To these we add two additional baselines: $(i)$ CoT refers to sampling $m$ predictions from  $\lbsfunc(x ; \theta^{(0)})$, and $(ii)$ TextGrad refers to first performing $T$ steps of prompt optimization and then sampling $m$ predictions from $\lbsfunc(x ; \theta^{(T)})$. 
% To ensure a fair comparison against \mcmcalg, we equate $m$ across all methods and use the same value of $T$ for both TextGrad and \mcmcalg. 
Both TextGrad and MHLP require a one-time initial fixed cost incurred by prompt optimization and MCMC, respectively, and we use the same value of $T$ for both. All methods use the same number $m$ of $\lbsfunc$ calls during inference to ensure a fair comparison from a computational perspective.

% We note that all methods require the same number of $\lbsfunc$ calls at inference time, though both TextGrad and MHLP require a one-time initial fixed cost incurred by prompt optimization and MCMC, respectively.

We reiterate \autoref{sec:method} in highlighting that we follow the common pipeline for quantifying uncertainty in two steps: $(i)$ generate a diverse answer set $y^{(1)}, \ldots, y^{(m)}$ and $(ii)$ summarize them into an uncertainty score. Because ours is a method for step $(i)$, our baselines are methods designed specifically to do the same. This means we omit direct comparison to means of performing step $(ii)$ such as semantic entropy \citep{kuhn2023semantic} and other methods described in \autoref{sec:related}. Although these are also UQ methods, they are orthogonal to our approach, and can be straightforwardly combined (for an example, see \appref{app:semantic_entropy}). For direct comparison, all experiments in this section use confidence  or semantic confidence (described below) as the means of summarizing the uncertainty in every set of answers.

\textbf{Datasets}\quad We evaluate both predictive performance and model calibration on AIME 2024 \citep{aimedata}, SimpleQA \citep{wei2024measuring}, and QASPER \citep{qasper}, representing question-answering tasks that are closed-form, free-form, and free-form with context, respectively. We randomly select and fix 100 samples from each of SimpleQA and QASPER for all experiments and use all 30 available samples from AIME 2024.
Notably, QASPER includes contextless questions, which are explicitly marked as unanswerable. We use these instances to assess our method's ability to detect insufficient information and abstain from answering. See \appref{app:experiment} for further dataset details.

\begin{wrapfigure}[21]{r}{0.47\textwidth}
\vspace{-10pt}
    \centering
    \scriptsize
    \begin{minipage}{\linewidth}
        \captionof{table}{Accuracy (\%) across datasets}\label{table:acc}
                \setlength{\tabcolsep}{3pt} % Default value: 6pt
        \begin{tabular}{lccc}
            \hline
            Method & AIME & SimpleQA & QASPER \\
            \hline
            Paraphrasing & $12.6 \pm 0.7$ & $43.7 \pm 0.5$ & $43.7 \pm 1.3$ \\
            System-Message & $7.2 \pm 0.7$ & $\mathbf{47.3 \pm 0.7}$ & $\mathbf{59.7 \pm 0.6}$ \\
            CoT & $9.0 \pm 1.4$ & $\mathbf{47.8 \pm 0.6}$ & $56.5 \pm 0.8$ \\
            TextGrad & $11.9 \pm 0.9$ & $46.6 \pm 0.5$ & $58.8 \pm 1.0$ \\
            \mcmcalg (Ours) & $\mathbf{15.0 \pm 0.7}$ & $\mathbf{48.6 \pm 0.6}$ & $\mathbf{60.9 \pm 1.0}$ \\
            \hline
        \end{tabular}

        \vspace{1.5ex}

        \captionof{table}{ECE / SECE (\%) across datasets}\label{table:ece}
                \setlength{\tabcolsep}{3pt} % Default value: 6pt
        \begin{tabular}{lccc}
            \hline
            Method & AIME & SimpleQA & QASPER \\
            \hline
            Paraphrasing & $\mathbf{21.1 \pm 0.8}$ & $18.7 \pm 0.7$ & $28.5 \pm 1.1$ \\
            System-Message & $\mathbf{19.7 \pm 0.8}$ & $18.4 \pm 0.4$ & $23.9 \pm 0.9$ \\
            CoT & $31.5 \pm 1.4$ & $18.0 \pm 0.6$ & $26.2 \pm 0.67$ \\
            TextGrad & $27.4 \pm 1.6$ & $17.7 \pm 1.0$ & $21.6 \pm 1.2$ \\
            \mcmcalg (Ours) & $22.0 \pm 1.0$ & $\mathbf{15.4 \pm 0.6}$ & $\mathbf{17.7 \pm 1.1}$ \\
            \hline
        \end{tabular}
        \captionof{table}{Abstention ROC AUC (\%)}\label{table:abst_auc}
                \setlength{\tabcolsep}{3pt} % Default value: 6pt
        \centering
        \begin{tabular}{lcc}
            \hline
            \multirow{2}{*}{Method} & \multicolumn{2}{c}{QASPER} \\
            \cline{2-3}
            & No context & Random context \\
            \hline
            Paraphrasing & $48.2 \pm 1.1$ & $62.1 \pm 1.6$ \\
            System-Message & $\mathbf{76.6 \pm 1.7}$ & $\mathbf{69.9 \pm 1.3}$ \\
            CoT & $75.6 \pm 1.1$ & $67.4 \pm 0.9$ \\
            TextGrad & $66.6 \pm 2.1$ & $67.4 \pm 0.9$ \\
            \mcmcalg (Ours) & $\mathbf{77.9 \pm 1.2}$ & $\mathbf{71.7 \pm 0.9}$ \\
            \hline
        \end{tabular}
    \end{minipage}
\end{wrapfigure}

In Tab. \ref{table:acc}, we report accuracy for all datasets using exact-match on closed-form datasets and an LLM judge \citep{zheng2023judging} to assess semantic correctness on free-form datasets. In Tab. \ref{table:ece}, we report the expected calibration error (ECE) as a measure of model calibration \citep{naeini2015obtaining, guo2017calibration}. Additionally, for QASPER, we estimate abstention ability on two types of unanswerable questions: questions with no context, and those with a random context. We use the same confidence scores used to estimate calibration as an abstention metric and compute the ROC AUC of this score when used as a classifier of answerability. Results are shown in Tab. \ref{table:abst_auc}. All results are averaged over 10 independent runs with standard errors to account for stochasticity.
% ; see Appendix \todo{Add} for more details.

\textbf{Semantic ECE}\quad
Standard ECE cannot be applied to open-ended tasks since it requires a confidence score, which is nontrivial to compute in general due to the variability of possible correct responses. To address this limitation, inspired by semantic entropy~\citep{kuhn2023semantic}, we propose an extension of ECE based on semantic clustering. Our metric, semantic ECE (SECE), uses these clusters to estimate model confidence over free-form outputs. Specifically, for each input $x_i$, we sample $m$ outputs: $y_i^{(1)}, \ldots, y_i^{(m)}$. We then query an LLM to group these outputs into semantic clusters. The empirical probability assigned by the model to each cluster is defined as the proportion of the generated samples in that cluster. The maximum of these probabilities is then taken as the model's \emph{semantic confidence} for input $x_i$. Finally, we use this value as the confidence for standard ECE computation, enabling estimation of model calibration for free-form outputs.

% \begin{wrapfigure}[9]{r}{0.45\textwidth}
% \vspace{-12pt}
%     \centering
%     \small
%     \begin{minipage}{\linewidth}
%         \captionof{table}{Abstention ROC AUC (\%)}\label{table:abst_auc}
%                 \setlength{\tabcolsep}{3pt} % Default value: 6pt
%         \begin{tabular}{lcc}
%             \hline
%             \multirow{2}{*}{Method} & \multicolumn{2}{c}{QASPER} \\
%             \cline{2-3}
%             & No context & Random context \\
%             \hline
%             Paraphrasing & $48.2 \pm 1.1$ & $62.1 \pm 1.6$ \\
%             System-Message & $\mathbf{76.6 \pm 1.7}$ & $\mathbf{69.9 \pm 1.3}$ \\
%             CoT & $75.6 \pm 1.1$ & $67.4 \pm 0.9$ \\
%             TextGrad & $66.6 \pm 2.1$ & $67.4 \pm 0.9$ \\
%             \mcmcalg (Ours) & $\mathbf{77.9 \pm 1.2}$ & $\mathbf{71.7 \pm 0.9}$ \\
%             \hline
%         \end{tabular}
%     \end{minipage}
% \end{wrapfigure}

\textbf{Discussion}\quad Across tasks, \mcmcalg is the only method to consistently outperform the rest. It only trails in calibration (ECE) on AIME, but its accuracy exceeds the two best-calibrated methods by a substantial margin. We hypothesize this outperformance is due to the high-posterior-valued samples of $\theta$ generated by \mcmcalg; it effectively performs stochastic prompt optimization, incorporating quantitative performance into its accept/reject decisions. In contrast, TextGrad alone has no accept/reject scheme and thus ``always accepts'', leading to the inclusion of potentially less useful changes to the initial prompt. For qualitative examples and diagnostics of accept/reject decisions, see \appref{app:experiment}.

\subsection{Conformal factuality with \mcmcalg}\label{sec:conformal}

\textbf{Background}\quad Conformal factuality \citep{mohri2024factuality} is a method for providing statistical guarantees on the correctness of LLM-generated answers to open-ended questions based on conformal prediction (CP) \citep{vovk2005algorithmic, shafer2008tutorial}. Generally, CP techniques use a small set of $n$ labeled datapoints to calibrate a prediction threshold. %Conformal factuality extends CP from classification or regression tasks to the domain of open-ended question-answering. 
In conformal factuality, given a question $x$, an LLM generates an answer $y$ which is broken into a set of distinct claims $\{c_1, \dots, c_\ell\}$. Each claim is assigned a factuality score $\importancefunc(c; \theta)$---generated by an \lbs---with larger values indicating increased confidence that $c$ is a factual claim. Then, after using CP to calibrate a threshold $\lambda$, claims with $\importancefunc(c; \theta)<\lambda$ are filtered out, such that only high-confidence claims are returned in the final answer $\hat{y}$.
CP guarantees that $\hat y$ contains only factual claims with high probability,
\begin{equation}\label{eq:conformal_factuality}
    1 - \alpha \leq \mathbb{P}[c \text{ is factual } \forall \ c\in\hat y] \leq 1 - \alpha + \frac{1}{n+1},
\end{equation}
where the error rate $\alpha$ is user-defined.
The quality of final answers can be gauged through the fraction of claims which are retained, since longer answers with more claims are more useful.\footnote{Filtering out all claims guarantees that $\hat{y}$ does not contain false claims, but does not give a useful answer.} Better calibrated expressions of confidence through $\importancefunc(c; \theta)$ improve claim retention, and since \mcmcalg enables better calibration, we can use it to design a better factuality score.

\textbf{Baseline (GPT-4 frequency scoring)}\quad The best performing option for $\importancefunc(c; \theta)$ from \citet{mohri2024factuality} is frequency scoring. Five alternative answers $y^{(p)}$ are generated for the same question $x$ from GPT-4 \citep{Achiam2023GPT4TR} using unit temperature and a manually crafted prompt $\theta$. For each claim in the original answer $y$, the number of times it appears across the $y^{(p)}$, i.e.\ its self-consistency \citep{wang2023selfconsistency, manakul2023selfcheckgpt}, is used as the score $\importancefunc(c; \theta)$.

\textbf{Our method (\mcmcalg frequency scoring)}\quad Like GPT-4 frequency scoring, our method estimates a claim's importance based on its frequency across alternative generations. However, instead of generating with a single fixed prompt, we produce diverse alternatives by sampling different $\theta$ via \mcmcalg with zero temperature. Notably, in the factuality context, ground truth outputs $\{y_1, \ldots, y_n\}$ are unavailable, so the unnormalized posterior $\prior\likelihood$ is unavailable. To surmount this obstacle, we replace the unnormalized probability mass with a surrogate
\begin{equation}
    g(\theta) = \mathbb{E}_{p(y' \mid x, \theta)}[ \tfrac{1}{\vert y'\vert}\textstyle\sum_{c\in y'} \importancefunc(c; \theta)],
\end{equation}
which we estimate stochastically when running \autoref{alg:metropolis_hastings}. One alternative answer $y^{(p)}$ is generated per sampled prompt. The ability to sample from this surrogate distribution underscores \mcmcalg's versatility in situations beyond conventional Bayesian inference.

\textbf{Dataset}\quad We use FactScore \citep{min2023factscore}, which is widely adopted for factuality tests of LLMs. Following \citet{mohri2024factuality}, we focus on ``person'' entities from the biography generation subset and extract subclaims from the generated biographies using the same extraction method across runs. We also follow \citet{mohri2024factuality} in using 50 samples for the calibration/test sets and performing 1000 random splits of calibration and test data for each $\alpha$ value. 

\textbf{Implementation}\quad We initialize both scoring methods with the same prompt. For MHLP, we perform sampling using a separate set of 100 samples from FactScore and obtain five prompt samples. Since there is no ground-truth answer in the open-ended QA setting, factuality is determined by decomposing answers into claims (as in \citet{mohri2024factuality}) and annotating them using a GPT web search tool.  
We use GPT-4 for answer generation, and GPT-4o-mini for claim generation, factuality annotation, frequency scoring, and MHLP proposals. See \appref{app:experiment} for more details.

\begin{figure}[t]
  \centering
  \begin{subfigure}[b]{0.48\textwidth}
    \includegraphics[width=0.9\textwidth]{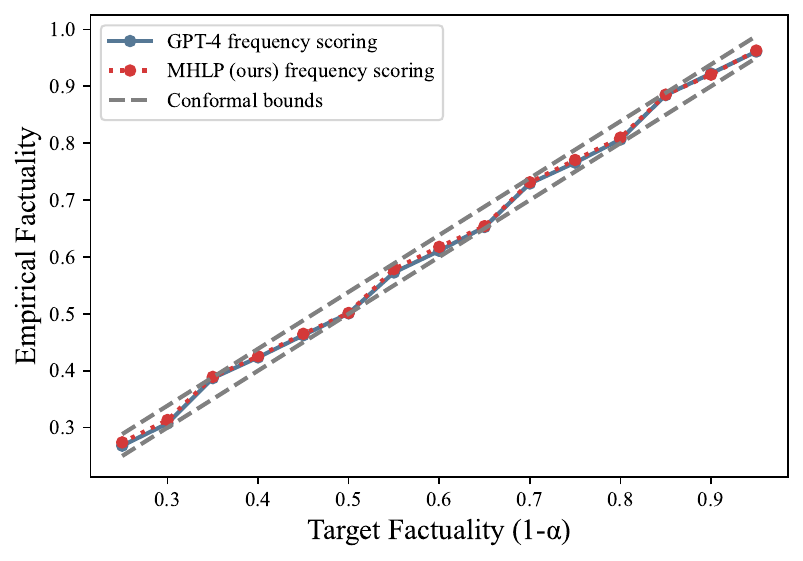}
    \caption{\footnotesize{Empirical factuality vs. Target factuality $1-\alpha$}}
    \label{fig:coverage}
  \end{subfigure}
  \hfill
  \begin{subfigure}[b]{0.48\textwidth}
    \includegraphics[width=0.9\textwidth]{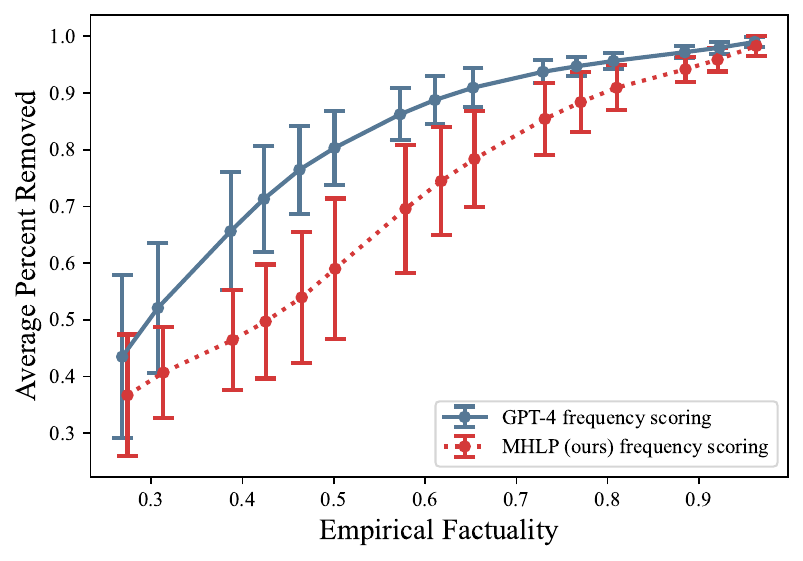}
    \caption{\footnotesize{Average removal rate vs. Empirical factuality}}
    \label{fig:removal}
  \end{subfigure}
  \caption{\footnotesize{Comparison of conformal factuality for frequency scoring with a fixed prompt \citep{mohri2024factuality}, and with prompts sampled through \mcmcalg. (a) The empirical factuality achieved in practice is consistently within the bounds guaranteed by \autoref{eq:conformal_factuality}. (b) \mcmcalg achieves the same level of empirical factuality as frequency scoring but removes fewer claims, indicating better calibrated confidence.}}
  \label{fig:conformal_factuality}
  \vspace{-15pt}
\end{figure}

\textbf{Results}\quad
First, we verify that both scoring methods achieve the target coverage from \autoref{eq:conformal_factuality}: \autoref{fig:coverage} shows that empirical factuality remains within the conformal bounds across all values of $\alpha$. \autoref{fig:removal} compares the removal rate, with error bars showing the standard deviation of the average removal rate across the 1000 data splits. Our method consistently achieves lower removal, showing that MHLP scoring provides a better uncertainty estimation of the factuality of LLM outputs.\footnote{The GPT-4 frequency scoring method shows slightly higher removal than reported by \citet{mohri2024factuality}, likely due to our use of a stricter web search–based factuality annotator.}

%% file: sections/5-related.tex
LLMs are applicable to a wide range of tasks and settings, which makes UQ inherently ambiguous--there is no single, well-defined quantity that UQ aims to approximate. Although our main setting of interest is where we have access to a pre-trained model and no fine-tuning is performed, we note that popular methods for UQ in deep learning, such as ensembles \citep{lakshminarayanan2017simple} and Laplace approximations \citep{ritter2018, kristiadi2020being, daxberger2021laplace}, have been successfully ported over for UQ when fine-tuning LLMs \citep{wang2023lora, yang2024bayesian}. Within our setting of interest, some approaches estimate uncertainty by analyzing the variability in outputs generated by an LLM given the same input \citep{kuhn2023semantic, lin2024generating, grewal2024improving, wang2024subjective, qiu2024semantic, nikitin2024kernel}, others do so by perturbing or modifying the input itself (e.g.\ by paraphrasing) \citep{hou2024decomposing, gao2024spuq, abbasi2024believe, zhang2024understanding, zhao2024knowing, feng2025rethinking}, and still others rely on directly asking the model to express its own confidence \citep{kadavath2022language, yang2024verbalized}. Unlike all these methods, we aim to quantify the uncertainty associated with LLM prompts. Other methods for UQ within in-context learning tasks with LLMs have also leveraged Bayesian ideas \citep{ling2024uncertainty, jesson2024estimating, tonolini2024bayesian, feng2025bird}, but we highlight that these works differ greatly from ours in that they do not directly perform Bayesian inference over free-form text and, once again, they do not quantify uncertainty over prompts.

Lastly, we mention another line of work performing Metropolis-Hastings over text. Like ours, \citet{faria2024quest} use LLMs to construct a proposal distribution within the MH algorithm. We nonetheless highlight many differences with this work: their method is applied to machine translation and not to UQ, they do not perform Bayesian inference, and their proposal is completely different and does not rely on prompt optimization methods. Also, concurrently to our work, \citet{faria2025sample} build on top of \citet{faria2024quest} by applying their proposal to a Bayesian formulation of the alignment problem wherein aligned model answers are sampled directly using MCMC. Investigating crossover applications of our and their proposals are potential directions for future work.

%% file: sections/6-conclusion.tex
In this work, we propose Textual Bayes for quantifying uncertainty in \lbss. Our work represents a formalization of recent work conceptualizing \lbss as models whose parameters are their prompts. Textual Bayes furthers this framework by performing Bayesian inference on these parameters, thus blending cutting-edge models with a formal statistical framework for uncertainty quantification. To implement this framework, we propose \mcmcalgfull (\mcmcalg), a novel MCMC algorithm for free-form text which finds applications in Bayesian inference and beyond. We test these frameworks on several uncertainty quantification benchmarks and find that they consistently improve the frontier of accuracy and calibration. We also show that \mcmcalg can be adapted to a factuality-based objective, leading to more reliable factual claims as quantified by the setting of conformal factuality.

Although Textual Bayes and \mcmcalg post strong performance against baselines, there remain avenues for improvement. First, MCMC is costly; despite equivalent inference cost to leading baselines, Textual Bayes requires a one-time expensive application of \mcmcalg. This cost might be addressed, for example, by further engineering the underlying prompt optimization method or training a small language model specifically for the task of generating proposals. Second, like many practical applications of Bayesian inference, our method requires approximations, which will inevitably cause deviations from the true posterior. Third, our evaluations on free-form answering benchmarks require LLM-based clustering. These techniques, though fairly applied across methods, are imperfect and a stronger evaluation signal might be obtained with improved fine-tuning, prompt engineering, or human evaluation. Lastly, we expect future work to find broader applications for \mcmcalg beyond Bayesian inference. For example, we could use \mcmcalg to modulate the \emph{outputs} of \lbss in accordance with unnormalized functions quantifying objectives such as alignment or safety.

% \textbf{Ethics Statement} \quad Not necessary for us

\textbf{Reproducibility statement}
\quad Code for this repo is available at \href{https://github.com/layer6ai-labs/textual-bayes}{https://github.com/layer6ai-labs/textual-bayes}. Our method is described in full throughout \autoref{sec:method} and \appref{app:approximations}, with experimental details in \autoref{sec:results} and \appref{app:hypers}. The datasets we benchmark on are publicly available.

%% file: sections/A-method.tex
\subsection{Useful Approximations}
\label{app:approximations}
In general, MCMC can only be applied to Bayesian inference when the $g(\theta)$ is calculable, where $g(\theta)$ is defined by 
\begin{equation}
    g(\theta) = p(\theta)p(\dataset \mid \theta) = p(\theta)\prod_{i=1}^n p(y_i \mid x_i, \theta).
\end{equation}
In our context, certain terms in this equation are intractable and we instead estimate them stochastically.

\paragraph{Mini-batch estimates of $\prod_{i=1}^n p(y_i \mid x_i, \theta)$}
Computing the term $\prod_{i=1}^n p(y_i \mid x_i, \theta)$ requires evaluating $\lbsfunc(x_i)$ for all $i \in \{1, \ldots, n\}$. Training sets are often large enough to make running a full pass per MCMC step intractable. Instead, we use mini-batching. One way to mini-batch is to apply the MH adjustment of \citet{seita2018efficient}, which involves a different accept/reject step; however, to avoid complexity we use a simple stochastic estimate of $\prod_{i=1}^n p(y_i \mid x_i, \theta)$ instead. Using a batch size of $b$, we make the estimate
\begin{equation}
    \prod_{i=1}^n p(y_i \mid x_i, \theta)\approx \prod_{j=1}^b p(y_{i_j} \mid x_{i_j}, \theta)^\frac{n}{b},
\end{equation}
which is an unbiased estimate in log-space:
\begin{equation}
    \sum_{i=1}^n \log p(y_i \mid x_i, \theta) = n\E_{(x, y) \sim \text{Uniform}(\dataset)}[\log p(y \mid x, \theta)] \approx  \frac{n}{b}\sum_{j=1}^b \log p(y_{i_j} \mid x_{i_j}, \theta),
\end{equation}
where $\{(x_{i_1}, y_{i_1}), \ldots, (x_{i_b}, y_{i_b})\}$ is a mini-batch of training datapoints.
In experiments, we use a batch size of $b=1$.

\paragraph{Tempered posterior} A well-studied phenomenon in Bayesian deep learning is the \emph{cold posterior effect} wherein sampling from the ``tempered'' posterior $p_\tau(\theta \mid \dataset) \propto \likelihood^{1 / \tau} \prior$ with $0 < \tau < 1$ often results in better empirical performance than the standard Bayesian posterior (i.e., $\tau = 1$) \citep{wenzel2020good, aitchison2020statistical, fortuin2021bayesian, izmailov2021bayesian, noci2021disentangling, kapoor2022uncertainty, nabarro2022data}. Following this practice, we apply a temperature $\tau$, making the final estimate equal to 
\begin{equation}
    \prod_{i=1}^n p(y_i \mid x_i, \theta)^\frac{1}{\tau}\approx \prod_{j=1}^b p(y_{i_j} \mid x_{i_j}, \theta)^\frac{n}{\tau b}.
\end{equation}
For simplicity, we absorb the exponent into a single constant $\beta := \frac{n}{\tau b}$ and tune $\beta$ for performance. As per the hyperparameter details below, often $\beta < n/b$ is the most effective, indicating a \emph{hot posterior} effect in our case.

\paragraph{Monte Carlo estimates of $p(y_i \mid x_i, \theta)$ with surrogate models}
For a single LLM call $y = \llm(x; \theta)$ that outputs an answer directly, computing $\log p(y \mid x, \theta)$ is as simple as summing log-probabilities across every token in $y$.
However, complex \lbss that include intermediate outputs and reasoning involve sources of stochasticity that are not captured in log-probabilities associated with the tokens of $y$.

Let $z$ be a variable capturing all intermediate outputs in the computation of $y = \lbsfunc(x; \theta)$. This includes internal LLM calls and reasoning text. We can express the generative process of sampling $y$ as
\begin{equation}
    z \sim p(z \mid x, \theta), 
    \hspace{40pt}
    y \sim p(y \mid z, x, \theta).
\end{equation}
Then the probability $p(y \mid x, \theta)$ required for MHLP would be computed as
\begin{equation}
    p(y \mid x, \theta) = \sum_z p(y \mid z, x, \theta)p(z \mid x, \theta) = \E_{z \sim p(z \mid x, \theta)}\left[p(y \mid z, x, \theta)\right].
\end{equation}
As suggested by the second equality, this intractable sum is again amenable to Monte Carlo estimates. We use a single sample of $z$ to estimate the likelihood in MHLP. We note that one alternative would be to remove all stochasticity from $z \sim p(z \mid x, \theta)$ by fixing a seed or setting the LLM temperature to 0 in the process of sampling $z$; however, this would remove necessary stochasticity from the final model result and alter the underlying model.

When using closed-source model providers, log-probabilities and thus the value of $p(y \mid z, x, \theta)$ itself is sometimes withheld. In this case, during MHLP we substitute the final LLM call in our \lbs with an open source model.

Lastly, we point out that all of the tricks above apply to computing probability masses for any \lbs, including the proposal density $q(\theta' \mid \theta)$ also required for MHLP.

\subsection{Updates with TextGrad}
\label{app:textgrad_update}

In our experiments, we implement $\promptopt$ as a TextGrad step \citep{yuksekgonul2025optimizing}. Given a model output $y_{\text{pred}} = \lbsfunc(x; \theta)$, we compute a loss string $\ell$ using a textual loss function $\tglossfunc$

\begin{equation}
\ell := \tglossfunc(x, y_\text{pred}, y) :=  \llm(x, y_\text{pred}, y; p).
\end{equation}

\begin{wrapfigure}[4]{r}{0.5\textwidth}
\vspace{-18pt}
\begin{minipage}{0.5\textwidth}
\begin{algorithm}[H]
\caption{TextGrad Update}\label{alg:textgrad_update}
\textbf{Require:} $\theta^{(t-1)}$\;
$\text{optimizer} \gets$ textgrad.Optimizer($\theta^{(t-1)}$)\;
$\text{optimizer.zero\_grad()}$\;
$y_{\text{pred}} \gets \lbsfunc(x; \theta^{(t-1)})$\;
$\ell \gets \tglossfunc(x, y, y_{\text{pred}})$\;
$\ell.\tgbackward()$\;
$\theta' \gets \tgstep()$\;
\KwRet{$\theta'$}\;
\end{algorithm}
\end{minipage}
\vspace{-2em}
\end{wrapfigure}

Where $p$ is a dataset-specific prompt describing how to evaluate a given model output, such as 
\begin{lstlisting}[language=,basicstyle=\ttfamily\tiny,
                   keywordstyle=,stringstyle=,commentstyle=,
                   escapeinside={(*@}{@*)}]
  You will be given a question related to scientific research papers and an answer attempted by a
  language model. Evaluate the attempted answer. Be smart, logical, and very critical.
  Do not solve the question. Just provide concise feedback.

  Question: {(*@ $x$ @*)}
  Attempted answer: {(*@ $y_\text{pred}$ @*)}
  True answer: {(*@ $y$ @*)}
\end{lstlisting}

TextGrad implements an autograd-style wrapper for all textual variables. This wrapper provides the method $\ell.\tgbackward()$, which ``backpropagates'' variable-wise feedback from the evaluation $\ell$. The TextGrad $\tgstep()$ method then incorporates this feedback to build a new parameter set $\theta'$. Pseudocode is given in \autoref{alg:textgrad_update}.

To run Metropolis-Hastings (\autoref{alg:metropolis_hastings}), we also need to compute the proposal density value $q(\theta', \mid \theta)$, where in our case $\theta' = \promptopt(\theta)$. Using the approximations described in \appref{app:approximations}, we need only compute logits from the final call $\tgstep$, and so we always compute this step with an open-source LLM (Llama-3.1-Nemotron-70B-Instruct-HF \citep{wang2024helpsteer2,bercovich2025llama}).

%% file: sections/B-experiment.tex
\subsection{Dataset Details}

AIME~\citep{aimedata}, released under the MIT license, contains problems from the American Invitational Mathematics Examination (AIME)—a prestigious high school competition known for its challenging mathematical questions. Each answer is an integer. The exam consists of 29 to 30 questions per year. For evaluation, we used the 2024 exam, which was not included in GPT’s training data.

SimpleQA~\citep{wei2024measuring}, released under the MIT license, is a benchmark that evaluates the ability of LLMs to answer short, fact-seeking questions. It covers a wide range of topics, including science, history, geography, history, politics, etc. Both its questions and answers are short and direct. In our experiments, we evaluated the models on a subset of 100 examples from the dataset.

QASPER~\citep{qasper}, released under the CC-BY-4.0 license, is a free-form question-answering dataset focused on scientific research papers. It contains 5,049 questions across 1,585 papers in the field of Natural Language Processing. Each question is based on the content of a specific paper. In our experiments, we provided the model with a passage from the paper that contains the answer (i.e., the context), and then posed the question for it to answer using that context. We evaluated our model on 100 samples from this dataset under two different scenarios. In the first scenario, the context was entirely missing for 35 of the samples. In the second, 33 samples were provided with randomly selected context \citep{qasper_abstain} that did not contain the correct answer. In both cases, the model was expected to abstain from answering.

\subsection{Hyperparameters}\label{app:hypers}

In the following experiments, we use the OpenAI API for calls to GPT-4o-mini and GPT-4o. As our surrogate model for probability mass estimates (see \appref{app:approximations}), we use Llama-3.1-Nemotron-70B-Instruct-HF \citep{wang2024helpsteer2,bercovich2025llama} through the Together AI API. 

For all LLM calls we use a temperature of 1. We ensure $m=10$ final answers are sampled for each method. For the Chain-of-Thought baseline, we used the initial prompt and sampled 10 answers, then aggregated the resulting answers. For TextGrad, we use the sample initial prompt but run TextGrad for a given number of steps before sampling 10 answers from the final prompt. For MCMC, we sample 10 individual prompts from a single MCMC chain and sample 1 answer from each. We tune the MHLP parameter $\beta$ (see \appref{app:method}) separately for each dataset. GPT-4o was employed for clustering and LLM-based evaluation. Further hyperparameter details are shown below and in our code (see especially the config files \texttt{qasper.yaml}, \texttt{aime.yaml}, and \texttt{simpleqa.yaml}).

\begin{table}[h!]
\centering
\caption{Hyperparameters used for each dataset and method}
\begin{tabular}{lllcccc}
\toprule
\textbf{Dataset} & \textbf{Method} & \textbf{Model} & \textbf{Steps $(T)$} & \boldmath$\beta$ & \textbf{Burn-in $(d)$} & \textbf{Thinning $(h)$} \\
\midrule
\multirow{3}{*}{AIME} 
& Chain-of-Thought & \multirow{3}{*}{GPT-4o}        & 0  & --  & -- & -- \\
& TextGrad         &                                & 60 & --  & -- & -- \\
& MHLP             &                                & 60 & 10  & 6  & 6 \\
\midrule
\multirow{3}{*}{SimpleQA} 
& Chain-of-Thought & \multirow{3}{*}{GPT-4o}        & 0  & --  & -- & -- \\
& TextGrad         &                                & 60 & --  & -- & -- \\
& MHLP             &                                & 60 & 100 & 6  & 6 \\
\midrule
\multirow{3}{*}{QASPER} 
& Chain-of-Thought & \multirow{3}{*}{GPT-4o-mini}   & 0  & --  & -- & -- \\
& TextGrad         &                                & 20 & --  & -- & -- \\
& MHLP             &                                & 20 & 100 & 2  & 2 \\
\bottomrule
\end{tabular}
\end{table}

For all methods, we fix a string at the end of the prompt describing standardized formatting instructions for the model's final answer. We extract this answer and evaluate likelihoods $p(y \mid z, x, \theta)$ only on this value, relegating any reasoning beforehand to the $z$ variable (see \appref{app:approximations}).

\subsubsection{Baselines}
\label{app:baseline}

We adapted the perturber baselines from SPUQ~\citep{gao2024spuq}, specifically selecting the Paraphrasing and System Message perturbers for comparison. For all runs, we used GPT-4o-mini and GPT-4o for a fair comparison. Our implementation differs from the original in several details:

\textbf{Paraphrasing}: Rather than using a single LLM call with JSON formatting to produce all paraphrases, we made separate LLM calls for each paraphrase (to avoid invalid JSON outputs from the LLM with the original prompt). We used the following prompt:

\begin{lstlisting}[language=,basicstyle=\ttfamily\tiny,keywordstyle=,stringstyle=,commentstyle=]
Suggest a way to paraphrase the text in triple quotes above.
If the original text is a question, please make sure that your answer is also a question.
If the original text has answer options, please make sure your answer also has those options in the same order.
Answer should ONLY be the paraphrase and nothing else.
\end{lstlisting}

\textbf{System Message}: Instead of sampling with replacement from the available prompts, we expanded the set of system prompts and sampled without replacement. We appended these system prompts to the beginning of the message chain to preserve any existing system prompts. This was crucial for maintaining the output format required by the evaluator (e.g., answers ending with \texttt{Answer: <THE ANSWER>}). The set of system prompts used was:

\begin{lstlisting}[basicstyle=\ttfamily\tiny]
"you are a helpful assistant"
"you are a question-answering assistant"
"you are a nice assistant"
"You are an AI support tool."
"You are a friendly helper."
"You are here to assist users."
"You provide useful answers."
"You are a kind AI agent."
"You offer good information."
"You are a smart assistant."
"You help with many tasks."
"You are a reliable AI."
"You give clear responses."
"You are an able assistant."
"You try to be useful."
"You are a positive AI."
"You guide users well."
"You are an adept helper."
"You simplify complex things."
"You are a virtual guide."
"You aim to be accurate."
\end{lstlisting}

\subsubsection{Conformal Factuality}

\paragraph{Factuality annotation}
Since there is no ground-truth output for the biography generation task, we assess the factuality of each generated answer by verifying its atomic sub-claims via web search. We use the GPT web search tool API\footnote{\url{https://platform.openai.com/docs/guides/tools-web-search?api-mode=responses}}, which allows the model to retrieve external evidence before making a judgment. Each sub-claim is labeled as factual (1) or not (0) based on the retrieved information.
We call the API as follows:
\begin{lstlisting}
response = GPT_client.responses.create(
    model="gpt-4o-mini",
    tools=[
        {
            "type": "web_search_preview",
            "search_context_size": "low"
        }
    ],
    input=prompt,
)
response_content = response.output_text
\end{lstlisting}

\paragraph{Model and Prompt Setup}
We use GPT-4 for base biography generation and GPT-4o-mini for claim decomposition, factuality annotation, and frequency-based entailment scoring. Both the baseline frequency scoring and MHLP initialization use the same default system prompt:
\texttt{"You are a helpful assistant. Write a bio for people."}
For frequency scoring, we generate five alternative answers using this prompt. All prompts are listed in Table~\ref{tab:prompts}.

\begin{table*}[ht!]
  \centering
  \begin{tabular}{p{15cm} }
    \small{\textbf{Subclaim Separator}} \\
    \scriptsize{\texttt{Please breakdown the following input into a set of small, independent claims (make sure not to add any information), and return the output as a jsonl, where each line is {subclaim:[CLAIM], gpt-score:[CONF]}.{\textbackslash n} The confidence score [CONF] should represent your confidence in the claim, where a 1 is obvious facts and results like `The earth is round' and `1+1=2'. A 0 is for claims that are very obscure or difficult for anyone to know, like the birthdays of non-notable people. If the input is short, it is fine to only return 1 claim. The input is: }}  \\
    \small{\textbf{Frequency scoring}} \\
    \scriptsize{\texttt{You will get a list of claims and piece of text. For each claim, score whether the text supports, contradicts, or is unrelated to the claim. Directly return a jsonl, where each line is \{"id":[CLAIM\_ID], "score":[SCORE]\}. Directly return the jsonl with no explanation or other formatting. For the [SCORE], return 1 for supports, -1 for contradicts, and 0 for unrelated. The claims are:\textcolor{gray}{\textbackslash n}\{claim\_string\}\textcolor{gray}{\textbackslash n\textbackslash n}The text is:\textcolor{gray}{\textbackslash n}\{output\}}}  \\ \\
    \small{\textbf{Factuality Annotation}} \\
    \scriptsize{\texttt{Please verify if each of these claims is factual.{\textbackslash n}Claims:{\textbackslash n}[claims\_text]{\textbackslash n}Return your answer as a JSON array, where each element is an object with these keys:
    \{"subclaim": "[CLAIM]", "factual": 1 or 0, "source": "source or explanation"\}{\textbackslash n}
    Format your response as a valid JSON array only, with no additional text or formatting.{\textbackslash n}
    Example:{\textbackslash n}
    [{\textbackslash n}
    \{"subclaim": "claim 1", "factual": 1, "source": "source"\},{\textbackslash n}
    \{"subclaim": "claim 2", "factual": 0, "source": "source"\}{\textbackslash n}
    ]{\textbackslash n}
    }}  \\
    
  \end{tabular}
  \caption{Prompts for sub-claim separator, frequency scoring, and factuality annotation. Note both sub-claim separator and frequency scoring prompts are the same as used in~\citep{mohri2024factuality}} 
  \label{tab:prompts}
\end{table*}

\paragraph{Hyperparameter}
We run a single Metropolis-Hastings chain with $T=20$ total steps, a burn-in of $d=4$, and a thinning interval of $h=4$, resulting in $m=4$ sampled prompts. Together with the initial prompt, we obtain 5 prompts in total, which are used to compute frequency scores.

\subsection{Examples}
In this section, we explore some examples of how the algorithm runs.

First, for SimpleQA and QASPER, we exhibit several example questions and answers comparing results from Textual Bayes to TextGrad. We show how the 10 answers sampled by each method are clustered, and the number of answers that fall into each cluster. Overall, we see that Textual Bayes's confidence levels are better calibrated to the model's correctness.

For AIME, we explore the algorithm's acceptance rate and individual accept/reject decisions over time.

\subsubsection{SimpleQA}

The following examples are selected from the SimpleQA dataset. The second example represents a case where the LLM appears truly to not know the answer; our method quantifies uncertainty better by expressing much lower confidence (40\%) than the TextGrad baseline.

\begin{lstlisting}[language=,basicstyle=\ttfamily\footnotesize, breaklines=true, columns=fullflexible, escapeinside=||, keywordstyle=\color{blue}, commentstyle=\color{gray}, stringstyle=\color{red}, showstringspaces=false]
|\textcolor{green}{Question:}| According to Medland, Sarah E.; Loesch, Danuta Z.; Mdzewski, Bogdan; Zhu, Gu; Montgomery, Grant W.; Martin, Nicholas G. (September 28, 2007), what chromosome location was identified as linked to the finger ridge counts of the ring, index, and middle fingers through multivariate linkage analysis?

|\textcolor{red}{Answer:}| 5q14.1
\end{lstlisting}
\begin{table}[h!]
\centering
\caption{Counts per semantic cluster for TextGrad and our method}
\begin{tabular}{lcc}
\toprule
\textbf{Semantic Cluster} & \textbf{TextGrad} & \textbf{Ours} \\
\midrule
5q14.1  & 3 & 7 \\
5q14.3  & 3 & 0 \\
5       & 1 & 1 \\
15q14   & 1 & 0 \\
21q22   & 1 & 0 \\
3q26    & 1 & 0 \\
5q13    & 0 & 1 \\
5q35    & 0 & 1 \\
\bottomrule
\end{tabular}
\end{table}

\clearpage

\begin{lstlisting}[language=,basicstyle=\ttfamily\footnotesize, breaklines=true, columns=fullflexible, escapeinside=||, keywordstyle=\color{blue}, commentstyle=\color{gray}, stringstyle=\color{red}, showstringspaces=false]
|\textcolor{green}{Question:}| What was the population of the town of Lesbury in Northumberland, England in the 2011 census?

|\textcolor{red}{Answer:}| 1007
\end{lstlisting}
\vspace{-8pt}
\begin{table}[h!]
\centering
\caption{Counts per semantic cluster for TextGrad and our method}
\begin{tabular}{lcc}
\toprule
\textbf{Semantic Cluster} & \textbf{TextGrad} & \textbf{Ours} \\
\midrule
1,154 & 7 & 4 \\
1,118 & 1 & 0 \\
1,057 & 1 & 0 \\
1,205 & 1 & 0 \\
1,264 & 0 & 1 \\
1,386 & 0 & 1 \\
1,122 & 0 & 1 \\
984   & 0 & 1 \\
1,187 & 0 & 1 \\
1,112 & 0 & 1 \\
\bottomrule
\end{tabular}
\vspace{-8pt}
\end{table}

\subsubsection{QASPER}
The following examples are selected from the QASPER dataset. Note that for the second example, the context given to the model is unrelated to the query, making the query unanswerable such that one would expect a well-calibrated LLM to express a high degree of uncertainty.

\begin{lstlisting}[language=,basicstyle=\ttfamily\footnotesize, breaklines=true, columns=fullflexible, escapeinside=||, keywordstyle=\color{blue}, commentstyle=\color{gray}, stringstyle=\color{red}, showstringspaces=false]
|\textcolor{blue}{Context:}|  We begin with a hate speech lexicon containing words and phrases identified by internet users as hate speech, compiled by Hatebase.org. Using the Twitter API we searched for tweets containing terms from the lexicon, resulting in a sample of tweets from 33,458 Twitter users. We extracted the time-line for each user, resulting in a set of 85.4 million tweets. From this corpus we then took a random sample of 25k tweets containing terms from the lexicon and had them manually coded by CrowdFlower (CF) workers. Workers were asked to label each tweet as one of three categories: hate speech, offensive but not hate speech, or neither offensive nor hate speech. They were provided with our definition along with a paragraph explaining it in further detail. Users were asked to think not just about the words appearing in a given tweet but about the context in which they were used. They were instructed that the presence of a particular word, however offensive, did not necessarily indicate a tweet is hate speech. Each tweet was coded by three or more people. The intercoder-agreement score provided by CF is 92%. We use the majority decision for each tweet to assign a label. Some tweets were not assigned labels as there was no majority class. This results in a sample of 24,802 labeled tweets.

|\textcolor{green}{Question:}| How long is their dataset?

|\textcolor{red}{Answer:}| 85400000
\end{lstlisting}

\begin{table}[h!]
\centering
\caption{Counts per semantic answer for TextGrad and our method}
\begin{tabular}{lcc}
\toprule
\textbf{Semantic Answer} & \textbf{TextGrad} & \textbf{Ours} \\
\midrule
85.4 million tweets & 1 & 6 \\
24,802 tweets       & 9 & 4 \\
\bottomrule
\end{tabular}
\end{table}

\begin{lstlisting}[language=,basicstyle=\ttfamily\footnotesize, breaklines=true, columns=fullflexible, escapeinside=||, keywordstyle=\color{blue}, commentstyle=\color{gray}, stringstyle=\color{red}, showstringspaces=false]
|\textcolor{blue}{Random Context:}| Figure FIGREF4 is the overview of the proposed method using character 3-gram embeddings (char3-MS-vec). As illustrated in this figure, our proposed method regards the sum of char3-MS-vec and the standard word embedding as an input of an RNN. In other words, let INLINEFORM0 be char INLINEFORM1 -MS-vec and we replace Equation with the following: DISPLAYFORM0

|\textcolor{green}{Question:}| Do they report results only on English data?

|\textcolor{red}{Answer:}| Unanswerable
\end{lstlisting}

\begin{table}[h!]
\centering
\caption{Counts per semantic answer for TextGrad and our method}
\begin{tabular}{lcc}
\toprule
\textbf{Semantic Answer} & \textbf{TextGrad} & \textbf{Ours} \\
\midrule
Unclear / not specified in context & 0 & 6 \\
Results are only on English data   & 0 & 1 \\
Results are not only on English    & 9 & 3 \\
Formatting error in answer         & 1 & 0 \\
\bottomrule
\end{tabular}
\end{table}

\subsubsection{AIME}

\begin{figure}[t]
    \centering
    \includegraphics[width=0.7\linewidth]{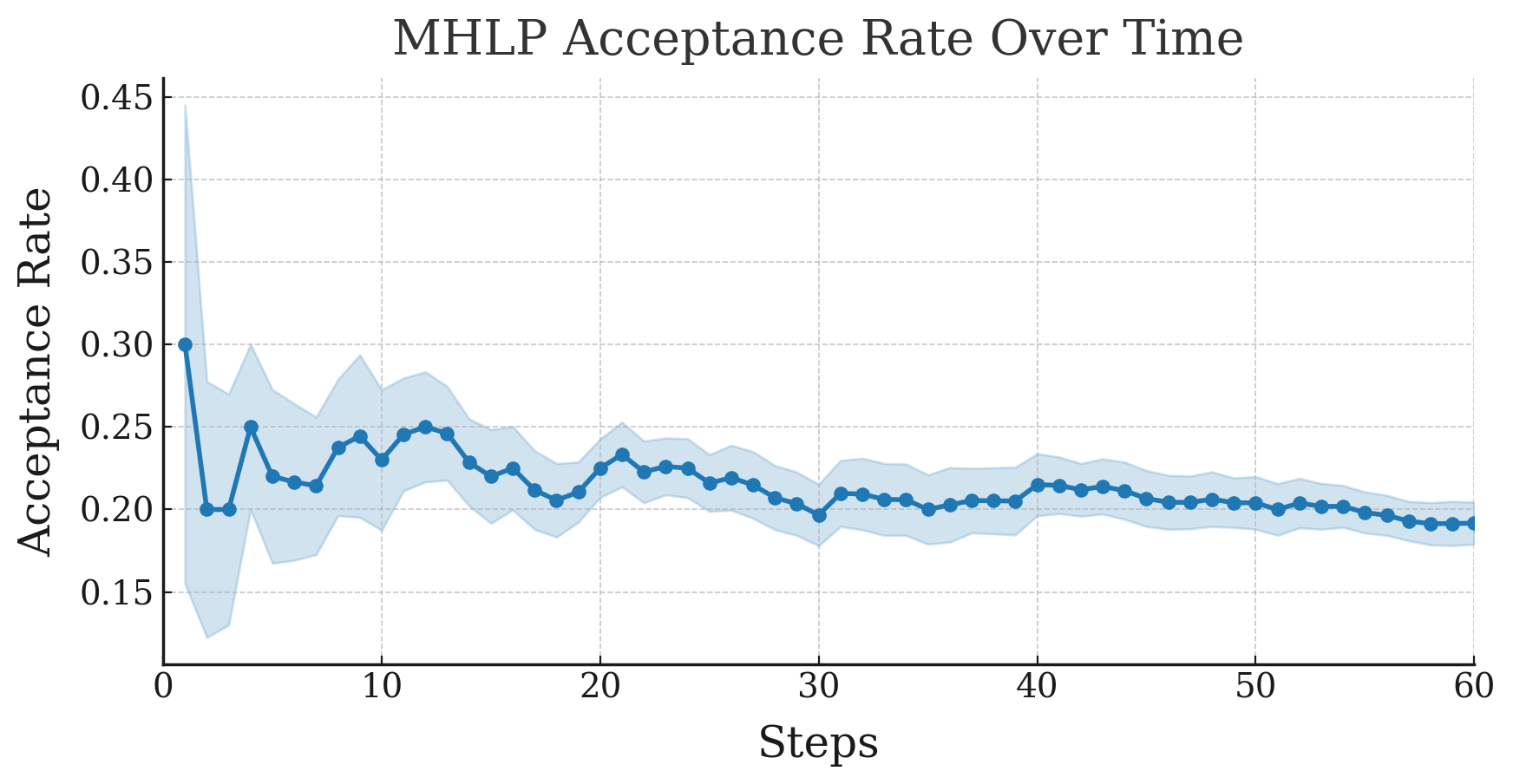}
    \caption{MHLP acceptance rate over time.}
    \label{fig:mhlp_accept}
\end{figure}

On the AIME dataset, we analyze the model's accept/reject decisions over time. In \autoref{fig:mhlp_accept}, we find that the model's acceptance rate over time closely matches the heuristic optimum of 0.234 prescribed by \citet{gelman1997weak}. \autoref{table:mhlp_trace} shows individual accept/reject decisions.

\begin{table}[h!]
\centering
\caption{AIME MHLP Accept/Reject Decisions}
\label{table:mhlp_trace}
\setlength{\tabcolsep}{3pt}
\renewcommand{\arraystretch}{1.15}
\small
\begin{tabularx}{\linewidth}{cYYc}
\toprule
\textbf{Step} & \textbf{Current prompt} & \textbf{Proposed prompt} & \textbf{Accepted (Y/N)} \\
\midrule
1 &
Answer the math question. Think step-by-step. &
Answer the math question by providing a step-by-step solution that explicitly considers all given conditions, such as divisibility requirements in terms of prime factor exponents. Employ appropriate combinatorial methods to account for problem constraints. Ensure accurate probability calculations based on correct combinatorial reasoning. Verify the simplification of your final answer, confirming the correct calculation of the greatest common divisor (GCD) if applicable. Present a clear, logical explanation for each step, and iteratively review and refine your solution to guarantee accuracy and clarity. &
Y \\
\midrule
20 &
Provide a concise, step-by-step solution to the math question, ensuring clear understanding by first restating the problem in your own words. Adopt a systematic approach, setting up relevant equations or congruences that directly address the problem's conditions. Verify each step's accuracy, correctly applying mathematical concepts such as modulo arithmetic, and maintain a logical flow by simplifying expressions where necessary. Ensure comprehensiveness by considering all possible solutions, and guarantee accuracy by cross-checking key calculations, ultimately leading to a precise final answer that encompasses all valid solution values. &
Provide a concise, step-by-step solution to the math question, ensuring clear understanding by first restating the problem in your own words. Apply relevant geometric principles and formulas directly related to the problem, avoiding unnecessary assumptions and concepts. Structure your response in a logical, connected manner, justifying each step. Verify calculations for accuracy, cross-checking key steps to guarantee a precise final answer that encompasses all valid solution values, based solely on the provided information and pertinent mathematical concepts. &
N \\
\midrule
40 &
Provide a concise, step-by-step solution to the math question, ensuring clear understanding by explicitly stating the problem's requirements and constraints. Justify your choice of coordinate systems and break down steps for deriving coordinates. Verify calculations, list and validate assumptions, and maintain a coherent logical flow with explanations for each decision. For geometric problems, explicitly calculate and present coordinates of critical intersection points and provide detailed, logically sound methods for calculating areas. Finally, re-evaluate your approach, cross-reference with alternative methods if possible, and verify the correctness of your answer to ensure accuracy and relevance. &
Provide a concise, step-by-step solution to the math question, ensuring clarity and accuracy. Analyze the problem's recursive definition (if applicable), explicitly describing the transformation of key elements (e.g., zeros, functions) at each step. Derive any formulas used for calculations from the given definitions, logically connecting each step. Include a detailed breakdown of intermediate calculations and explanations for each decision. Where applicable, utilize visual aids or specific examples to illustrate complex transformations. Finally, cross-verify your approach by considering alternative methods or perspectives, and validate your final answer to ensure accuracy and relevance. &
Y \\
\bottomrule
\end{tabularx}
\end{table}

\clearpage

\subsection{Semantic Entropy}
\label{app:semantic_entropy}

In the main text, we quantify uncertainty for our method and all baselines using confidence: the probability a model assigns to a given answer (or estimates thereof). Confidence is useful because it has a clear mathematical interpretation and can be used to assess calibration, but as outlined in \autoref{sec:related}, there a numerous other ways to compute uncertainty scores from \lbss. 

A popular uncertainty score among these is semantic entropy \citep{kuhn2023semantic}. In \autoref{table:semantic_entropy}, we check whether our performance is robust to alternate ways of estimating model uncertainty by using semantic entropy as an abstention score on the QASPER dataset, where unanswerable questions are those with the context removed. We find that the relative performances of methods in \autoref{sec:results} using confidence match those using semantic entropy.

\captionof{table}{QASPER - Abstention ROC AUC (\%) with Semantic Entropy}\label{table:semantic_entropy}
\setlength{\tabcolsep}{3pt} % Default value: 6pt
\centering
\begin{tabular}{lc}
    \hline
    Method & ROC \\
    \hline
    Paraphrasing+SE & $50.0 \pm 1.4$ \\
    System-Message+SE & $68.1 \pm 1.7$ \\
    CoT+SE & $71.3 \pm 1.8$ \\
    TextGrad+SE & $70.2 \pm 1.1$ \\
    MHLP+SE & $\mathbf{78.2 \pm 1.1}$ \\
    \hline
\end{tabular}

%% file: bib.bib
@inproceedings{
wang2024helpsteer2,
title={HelpSteer2-Preference: Complementing Ratings with Preferences},
author={Zhilin Wang and Alexander Bukharin and Olivier Delalleau and Daniel Egert and Gerald Shen and Jiaqi Zeng and Oleksii Kuchaiev and Yi Dong},
booktitle={The Thirteenth International Conference on Learning Representations},
year={2025},
}

@article{bercovich2025llama,
  title={Llama-nemotron: Efficient reasoning models},
  author={Bercovich, Akhiad and Levy, Itay and Golan, Izik and Dabbah, Mohammad and El-Yaniv, Ran and Puny, Omri and Galil, Ido and Moshe, Zach and Ronen, Tomer and Nabwani, Najeeb and others},
  journal={arXiv:2505.00949},
  year={2025}
}

@Inbook{chaturvedi2023opportunities,
author="Chaturvedi, Rijul
and Verma, Sanjeev",
title="Opportunities and Challenges of AI-Driven Customer Service",
bookTitle="Artificial Intelligence in Customer Service: The Next Frontier for Personalized Engagement",
year="2023",
publisher="Springer International Publishing",
pages="33--71",
isbn="978-3-031-33898-4",
doi="10.1007/978-3-031-33898-4_3",
}

@article{chen2021evaluating,
  title={Evaluating large language models trained on code},
author={Mark Chen and Jerry Tworek and Heewoo Jun and Qiming Yuan and Henrique Ponde de Oliveira Pinto and Jared Kaplan and Harri Edwards and Yuri Burda and Nicholas Joseph and Greg Brockman and Alex Ray and Raul Puri and Gretchen Krueger and Michael Petrov and Heidy Khlaaf and Girish Sastry and Pamela Mishkin and Brooke Chan and Scott Gray and Nick Ryder and Mikhail Pavlov and Alethea Power and Lukasz Kaiser and Mohammad Bavarian and Clemens Winter and Philippe Tillet and Felipe Petroski Such and Dave Cummings and Matthias Plappert and Fotios Chantzis and Elizabeth Barnes and Ariel Herbert-Voss and William Hebgen Guss and Alex Nichol and Alex Paino and Nikolas Tezak and Jie Tang and Igor Babuschkin and Suchir Balaji and Shantanu Jain and William Saunders and Christopher Hesse and Andrew N. Carr and Jan Leike and Josh Achiam and Vedant Misra and Evan Morikawa and Alec Radford and Matthew Knight and Miles Brundage and Mira Murati and Katie Mayer and Peter Welinder and Bob McGrew and Dario Amodei and Sam McCandlish and Ilya Sutskever and Wojciech Zaremba},
  journal={arXiv:2107.03374},
  year={2021}
}

@article{Achiam2023GPT4TR,
  title={{GPT-4 Technical Report}},
  author={Achiam, Josh and others},
  journal={arXiv:2303.08774},
  year={2023}
}

@article{faria2025sample,
  title={Sample, Don't Search: Rethinking Test-Time Alignment for Language Models},
  author={Faria, Gon{\c{c}}alo and Smith, Noah A},
  journal={arXiv:2504.03790},
  year={2025}
}

@inproceedings{min2023factscore,
    title = "{FActScore: Fine-grained Atomic Evaluation of Factual Precision in Long Form Text Generation}",
    author = "Min, Sewon  and
      Krishna, Kalpesh  and
      Lyu, Xinxi  and
      Lewis, Mike  and
      Yih, Wen-tau  and
      Koh, Pang  and
      Iyyer, Mohit  and
      Zettlemoyer, Luke  and
      Hajishirzi, Hannaneh",
    booktitle = "Proceedings of the 2023 Conference on Empirical Methods in Natural Language Processing",
    year = "2023",
    publisher = "Association for Computational Linguistics",
    doi = "10.18653/v1/2023.emnlp-main.741",
}

@article{wei2024measuring,
  title={Measuring short-form factuality in large language models},
  author={Wei, Jason and Karina, Nguyen and Chung, Hyung Won and Jiao, Yunxin Joy and Papay, Spencer and Glaese, Amelia and Schulman, John and Fedus, William},
  journal={arXiv:2411.04368},
  year={2024}
}

@article{kadavath2022language,
  title={Language models (mostly) know what they know},
  author={Kadavath, Saurav and Conerly, Tom and Askell, Amanda and Henighan, Tom and Drain, Dawn and Perez, Ethan and Schiefer, Nicholas and Hatfield-Dodds, Zac and DasSarma, Nova and Tran-Johnson, Eli and others},
  journal={arXiv:2207.05221},
  year={2022}
}

@inproceedings{
wang2023selfconsistency,
title={Self-Consistency Improves Chain of Thought Reasoning in Language Models},
author={Xuezhi Wang and Jason Wei and Dale Schuurmans and Quoc V Le and Ed H. Chi and Sharan Narang and Aakanksha Chowdhery and Denny Zhou},
booktitle={The Eleventh International Conference on Learning Representations },
year={2023},
}

@inproceedings{manakul2023selfcheckgpt,
    title = "{SelfCheckGPT: Zero-Resource Black-Box Hallucination Detection for Generative Large Language Models}",
    author = "Manakul, Potsawee  and
      Liusie, Adian  and
      Gales, Mark",
    booktitle = "Proceedings of the 2023 Conference on Empirical Methods in Natural Language Processing",
    year = "2023",
    publisher = "Association for Computational Linguistics",
    doi = "10.18653/v1/2023.emnlp-main.557",
    pages = "9004--9017",
}

@article{shafer2008tutorial,
  title={A Tutorial on Conformal Prediction.},
  author={Shafer, Glenn and Vovk, Vladimir},
  journal={Journal of Machine Learning Research},
  volume={9},
  number={3},
  year={2008}
}

@book{vovk2005algorithmic,
  title={Algorithmic Learning in a Random World},
  author={Vovk, Vladimir and Gammerman, Alexander and Shafer, Glenn},
  year={2005},
  publisher={Springer}
}

@InProceedings{mohri2024factuality,
  title = 	 {Language Models with Conformal Factuality Guarantees},
  author =       {Mohri, Christopher and Hashimoto, Tatsunori},
  booktitle = 	 {Proceedings of the 41st International Conference on Machine Learning},
  year = 	 {2024},
}

@inproceedings{wang2021codet5,
  title={CodeT5: Identifier-aware Unified Pre-trained Encoder-Decoder Models for Code Understanding and Generation},
  author={Wang, Yue and Wang, Weishi and Joty, Shafiq and Hoi, Steven CH},
  booktitle={Proceedings of the 2021 Conference on Empirical Methods in Natural Language Processing},
  pages={8696--8708},
  year={2021}
}

@article{boiko2023emergent,
  title={Emergent autonomous scientific research capabilities of large language models},
  author={Boiko, Daniil A and MacKnight, Robert and Gomes, Gabe},
  journal={arXiv:2304.05332},
  year={2023}
}

@article{schmidgall2025agent,
  title={{Agent Laboratory: Using LLM Agents as Research Assistants}},
  author={Schmidgall, Samuel and Su, Yusheng and Wang, Ze and Sun, Ximeng and Wu, Jialian and Yu, Xiaodong and Liu, Jiang and Liu, Zicheng and Barsoum, Emad},
  journal={arXiv:2501.04227},
  year={2025}
}

@article{yamada2025ai,
  title={{The AI Scientist-v2: Workshop-Level Automated Scientific Discovery via Agentic Tree Search}},
  author={Yamada, Yutaro and Lange, Robert Tjarko and Lu, Cong and Hu, Shengran and Lu, Chris and Foerster, Jakob and Clune, Jeff and Ha, David},
  journal={arXiv:2504.08066},
  year={2025}
}

@inproceedings{gomez2023confederacy,
  title={{A Confederacy of Models: a Comprehensive Evaluation of LLMs on Creative Writing}},
  author={G{\'o}mez-Rodr{\'\i}guez, Carlos and Williams, Paul},
  booktitle={Findings of the Association for Computational Linguistics: EMNLP 2023},
  pages={14504--14528},
  year={2023}
}

@article{wei2023jailbroken,
  title={{Jailbroken: How does LLM safety training fail?}},
  author={Wei, Alexander and Haghtalab, Nika and Steinhardt, Jacob},
  journal={Advances in Neural Information Processing Systems},
  volume={36},
  pages={80079--80110},
  year={2023}
}

@article{zou2023universal,
  title={Universal and transferable adversarial attacks on aligned language models},
  author={Zou, Andy and Wang, Zifan and Carlini, Nicholas and Nasr, Milad and Kolter, J Zico and Fredrikson, Matt},
  journal={arXiv:2307.15043},
  year={2023}
}

@inproceedings{yan2024backdooring,
  title={Backdooring Instruction-Tuned Large Language Models with Virtual Prompt Injection},
  author={Yan, Jun and Yadav, Vikas and Li, Shiyang and Chen, Lichang and Tang, Zheng and Wang, Hai and Srinivasan, Vijay and Ren, Xiang and Jin, Hongxia},
  booktitle={Proceedings of the 2024 Conference of the North American Chapter of the Association for Computational Linguistics: Human Language Technologies (Volume 1: Long Papers)},
  pages={6065--6086},
  year={2024}
}

@inproceedings{wei2022chain,
 author = {Wei, Jason and Wang, Xuezhi and Schuurmans, Dale and Bosma, Maarten and Ichter, Brian and Xia, Fei and Chi, Ed and Le, Quoc V and Zhou, Denny},
 booktitle = {Advances in Neural Information Processing Systems},
 pages = {24824--24837},
 title = {Chain-of-Thought Prompting Elicits Reasoning in Large Language Models},
 volume = {35},
 year = {2022}
}

@inproceedings{lewis2020retrieval,
 author = {Lewis, Patrick and Perez, Ethan and Piktus, Aleksandra and Petroni, Fabio and Karpukhin, Vladimir and Goyal, Naman and K\"{u}ttler, Heinrich and Lewis, Mike and Yih, Wen-tau and Rockt\"{a}schel, Tim and Riedel, Sebastian and Kiela, Douwe},
 booktitle = {Advances in Neural Information Processing Systems},
 pages = {9459--9474},
 title = {{Retrieval-Augmented Generation for Knowledge-Intensive NLP Tasks}},
 volume = {33},
 year = {2020}
}

@article{wooldridge1995intelligent,
  title={Intelligent agents: Theory and practice},
  author={Wooldridge, Michael and Jennings, Nicholas R},
  journal={The Knowledge Engineering Review},
  volume={10},
  number={2},
  pages={115--152},
  year={1995},
  publisher={Cambridge University Press}
}

@article{xi2025rise,
  title={The rise and potential of large language model based agents: A survey},
author={Zhiheng Xi and Wenxiang Chen and Xin Guo and Wei He and Yiwen Ding and Boyang Hong and Ming Zhang and Junzhe Wang and Senjie Jin and Enyu Zhou and Rui Zheng and Xiaoran Fan and Xiao Wang and Limao Xiong and Yuhao Zhou and Weiran Wang and Changhao Jiang and Yicheng Zou and Xiangyang Liu and Zhangyue Yin and Shihan Dou and Rongxiang Weng and Wensen Cheng and Qi Zhang and Wenjuan Qin and Yongyan Zheng and Xipeng Qiu and Xuanjing Huang and Tao Gui},
journal={Science China Information Sciences},
  volume={68},
  number={2},
  pages={121101},
  year={2025},
  publisher={Springer}
}

@inproceedings{
khattab2024dspy,
title={{DSP}y: Compiling Declarative Language Model Calls into State-of-the-Art Pipelines},
author={Omar Khattab and Arnav Singhvi and Paridhi Maheshwari and Zhiyuan Zhang and Keshav Santhanam and Sri Vardhamanan A and Saiful Haq and Ashutosh Sharma and Thomas T. Joshi and Hanna Moazam and Heather Miller and Matei Zaharia and Christopher Potts},
booktitle={The Twelfth International Conference on Learning Representations},
year={2024},
}

@inproceedings{madaan2023self,
 author = {Madaan, Aman and Tandon, Niket and Gupta, Prakhar and Hallinan, Skyler and Gao, Luyu and Wiegreffe, Sarah and Alon, Uri and Dziri, Nouha and Prabhumoye, Shrimai and Yang, Yiming and Gupta, Shashank and Majumder, Bodhisattwa Prasad and Hermann, Katherine and Welleck, Sean and Yazdanbakhsh, Amir and Clark, Peter},
 booktitle = {Advances in Neural Information Processing Systems},
 pages = {46534--46594},
 title = {{Self-Refine: Iterative Refinement with Self-Feedback}},
 volume = {36},
 year = {2023}
}

@inproceedings{zhugegptswarm,
  title={{GPTSwarm: Language Agents as Optimizable Graphs}},
  author={Zhuge, Mingchen and Wang, Wenyi and Kirsch, Louis and Faccio, Francesco and Khizbullin, Dmitrii and Schmidhuber, J{\"u}rgen},
  booktitle={Forty-first International Conference on Machine Learning},
  year={2024},
}

@article{xu2024hallucination,
  title={Hallucination is inevitable: An innate limitation of large language models},
  author={Xu, Ziwei and Jain, Sanjay and Kankanhalli, Mohan},
  journal={arXiv:2401.11817},
  year={2024}
}

@InProceedings{guo2017calibration,
  title = 	 {On Calibration of Modern Neural Networks},
  author =       {Chuan Guo and Geoff Pleiss and Yu Sun and Kilian Q. Weinberger},
  booktitle = 	 {Proceedings of the 34th International Conference on Machine Learning},
  pages = 	 {1321--1330},
  year = 	 {2017},
  volume = 	 {70},
}

@misc{aimedata,
  author       = {MAA},
  year = {2024},
  title        = {AIME Problems and Solutions},
  howpublished = {\url{https://artofproblemsolving.com/wiki/index.php/AIME_Problems_and_Solutions}},
}

@article{zheng2023judging,
  title={{Judging LLM-as-a-Judge with MT-Bench and Chatbot Arena}},
  author={Zheng, Lianmin and Chiang, Wei-Lin and Sheng, Ying and Zhuang, Siyuan and Wu, Zhanghao and Zhuang, Yonghao and Lin, Zi and Li, Zhuohan and Li, Dacheng and Xing, Eric and others},
  journal={Advances in Neural Information Processing Systems},
  volume={36},
  pages={46595--46623},
  year={2023}
}

@article{naeini2015obtaining, 
title={Obtaining Well Calibrated Probabilities Using Bayesian Binning}, 
volume={29}, 
DOI={10.1609/aaai.v29i1.9602},
number={1}, 
journal={Proceedings of the AAAI Conference on Artificial Intelligence},
author={Naeini, Mahdi Pakdaman and Cooper, Gregory and Hauskrecht, Milos}, 
year={2015}, 
}

@inproceedings{
aitchison2020statistical,
title={A statistical theory of cold posteriors in deep neural networks},
author={Laurence Aitchison},
booktitle={International Conference on Learning Representations},
year={2021},
}

@inproceedings{kapoor2022uncertainty,
 author = {Kapoor, Sanyam and Maddox, Wesley J and Izmailov, Pavel and Wilson, Andrew G},
 booktitle = {Advances in Neural Information Processing Systems},
 title = {On Uncertainty, Tempering, and Data Augmentation in Bayesian Classification},
 volume = {35},
 year = {2022}
}

@inproceedings{noci2021disentangling,
 author = {Noci, Lorenzo and Roth, Kevin and Bachmann, Gregor and Nowozin, Sebastian and Hofmann, Thomas},
 booktitle = {Advances in Neural Information Processing Systems},
 title = {Disentangling the Roles of Curation, Data-Augmentation and the Prior in the Cold Posterior Effect},
 volume = {34},
 year = {2021}
}

@inproceedings{nabarro2022data,
  title={{Data augmentation in Bayesian neural networks and the cold posterior effect}},
  author={Nabarro, Seth and Ganev, Stoil and Garriga-Alonso, Adri{\`a} and Fortuin, Vincent and van der Wilk, Mark and Aitchison, Laurence},
  booktitle={Uncertainty in Artificial Intelligence},
  pages={1434--1444},
  year={2022},
  organization={PMLR}
}

@InProceedings{izmailov2021bayesian,
  title = 	 {{What Are Bayesian Neural Network Posteriors Really Like?}},
  author =       {Izmailov, Pavel and Vikram, Sharad and Hoffman, Matthew D and Wilson, Andrew Gordon},
  booktitle = 	 {Proceedings of the 38th International Conference on Machine Learning},
  pages = 	 {4629--4640},
  year = 	 {2021},
  volume = 	 {139},
}

@inproceedings{zhou2022large,
  title={Large language models are human-level prompt engineers},
  author={Zhou, Yongchao and Muresanu, Andrei Ioan and Han, Ziwen and Paster, Keiran and Pitis, Silviu and Chan, Harris and Ba, Jimmy},
  booktitle={The Eleventh International Conference on Learning Representations},
  year={2022}
}

@inproceedings{cheng2024trace,
 author = {Cheng, Ching-An and Nie, Allen and Swaminathan, Adith},
 booktitle = {Advances in Neural Information Processing Systems},
 pages = {71596--71642},
 title = {{Trace is the Next AutoDiff: Generative Optimization with Rich Feedback, Execution Traces, and LLMs}},
 volume = {37},
 year = {2024}
}

@article{hu2024automated,
  title={Automated design of agentic systems},
  author={Hu, Shengran and Lu, Cong and Clune, Jeff},
  journal={arXiv:2408.08435},
  year={2024}
}

@article{yuksekgonul2025optimizing,
  title={{Optimizing generative AI by backpropagating language model feedback}},
  author={Yuksekgonul, Mert and Bianchi, Federico and Boen, Joseph and Liu, Sheng and Lu, Pan and Huang, Zhi and Guestrin, Carlos and Zou, James},
  journal={Nature},
  volume={639},
  pages={609--616},
  year={2025},
}

@inproceedings{maynez2020faithfulness,
  title={On Faithfulness and Factuality in Abstractive Summarization},
  author={Maynez, Joshua and Narayan, Shashi and Bohnet, Bernd and McDonald, Ryan},
  booktitle={Proceedings of the 58th Annual Meeting of the Association for Computational Linguistics},
  pages={1906--1919},
  year={2020}
}

@book{neal1996bayesian,
  title     = {Bayesian Learning for Neural Networks},
  author    = {Neal, Radford M.},
  year      = {1996},
  publisher = {Springer},
  series    = {Lecture Notes in Statistics},
  volume    = {118},
  doi       = {10.1007/978-1-4612-0745-0}
}

@article{saul1996mean,
  title={Mean field theory for sigmoid belief networks},
  author={Saul, Lawrence K and Jaakkola, Tommi and Jordan, Michael I},
  journal={Journal of Artificial Intelligence Research},
  volume={4},
  pages={61--76},
  year={1996}
}

@book{mackay2003information,
  title={Information Theory, Inference and Learning Algorithms},
  author={MacKay, David JC},
  year={2003},
  publisher={Cambridge University Press}
}

@book{bernardo2009bayesian,
  title={{Bayesian Theory}},
  author={Bernardo, Jos{\'e} M and Smith, Adrian FM},
  volume={405},
  year={2009},
  publisher={John Wiley \& Sons}
}

@InProceedings{welling2011bayesian,
  author =    {Max Welling and Yee Whye Teh},
  title =     {{Bayesian Learning via Stochastic Gradient Langevin Dynamics}},
  booktitle = {Proceedings of the 28th International Conference on Machine Learning},
  year =      {2011},
  isbn =      {978-1-4503-0619-5},
  pages=      {681--688},
}

@inproceedings{kingma2014auto,
  title={{Auto-encoding variational Bayes}},
  author={Kingma, Diederik P and Welling, Max},
  booktitle={International Conference on Learning Representations},
  year={2014}
}

@InProceedings{blundell2015weight,
  title = 	 {Weight Uncertainty in Neural Network},
  author = 	 {Blundell, Charles and Cornebise, Julien and Kavukcuoglu, Koray and Wierstra, Daan},
  booktitle = 	 {Proceedings of the 32nd International Conference on Machine Learning},
  pages = 	 {1613--1622},
  year = 	 {2015},
  volume = 	 {37},
}

@InProceedings{wenzel2020good,
  title = 	 {How Good is the {B}ayes Posterior in Deep Neural Networks Really?},
  author =       {Wenzel, Florian and Roth, Kevin and Veeling, Bastiaan and Swiatkowski, Jakub and Tran, Linh and Mandt, Stephan and Snoek, Jasper and Salimans, Tim and Jenatton, Rodolphe and Nowozin, Sebastian},
  booktitle = 	 {Proceedings of the 37th International Conference on Machine Learning},
  pages = 	 {10248--10259},
  year = 	 {2020},
  volume = 	 {119},
}

@inproceedings{
fortuin2021bayesian,
title={Bayesian Neural Network Priors Revisited},
author={Vincent Fortuin and Adri{\`a} Garriga-Alonso and Sebastian W. Ober and Florian Wenzel and Gunnar Ratsch and Richard E Turner and Mark van der Wilk and Laurence Aitchison},
booktitle={International Conference on Learning Representations},
year={2022},
}

@article{duane1987hybrid,
  title     = {Hybrid Monte Carlo},
  author    = {Duane, Simon and Kennedy, A. D. and Pendleton, B. J. and Roweth, Duncan},
  journal   = {Physics Letters B},
  volume    = {195},
  number    = {2},
  pages     = {216--222},
  year      = {1987},
}

@article{gelman1997weak,
  title={Weak convergence and optimal scaling of random walk Metropolis algorithms},
  author={Gelman, Andrew and Gilks, Walter R and Roberts, Gareth O},
  journal={The Annals of Applied Probability},
  volume={7},
  number={1},
  pages={110--120},
  year={1997},
  publisher={Institute of Mathematical Statistics}
}

@article{rosenthal2011optimal,
  title={{Optimal proposal distributions and adaptive MCMC}},
  author={Rosenthal, Jeffrey S},
  journal={Handbook of Markov Chain Monte Carlo},
  volume={4},
  number={10.1201},
  pages={93--111},
  year={2011},
  publisher={Chapman \& Hall/CRC Boca Raton, FL}
}

@article{robert2015metropolis,
  title={{The Metropolis-Hastings algorithm}},
  author={Robert, Christian P},
  journal={arXiv:1504.01896},
  year={2015}
}

@inproceedings{seita2018efficient,
  title={An efficient minibatch acceptance test for metropolis-hastings},
  author={Seita, Daniel and Pan, Xinlei and Chen, Haoyu and Canny, John},
  booktitle={Proceedings of the 27th International Joint Conference on Artificial Intelligence},
  pages={5359--5363},
  year={2018}
}

@inproceedings{qasper,
    title = "A Dataset of Information-Seeking Questions and Answers Anchored in Research Papers",
    author = "Dasigi, Pradeep  and
      Lo, Kyle  and
      Beltagy, Iz  and
      Cohan, Arman  and
      Smith, Noah A.  and
      Gardner, Matt",
    booktitle = "Proceedings of the 2021 Conference of the North American Chapter of the Association for Computational Linguistics: Human Language Technologies",
    year = "2021",
    doi = "10.18653/v1/2021.naacl-main.365",
}

@inproceedings{qasper_abstain,
    title = "Characterizing {LLM} Abstention Behavior in Science {QA} with Context Perturbations",
    author = "Wen, Bingbing  and
      Howe, Bill  and
      Wang, Lucy Lu",
    booktitle = "Findings of the Association for Computational Linguistics: EMNLP 2024",
    year = "2024",
    doi = "10.18653/v1/2024.findings-emnlp.197",
}

@inproceedings{kuhn2023semantic,
  title={Semantic uncertainty: Linguistic invariances for uncertainty estimation in natural language generation},
  author={Kuhn, Lorenz and Gal, Yarin and Farquhar, Sebastian},
  booktitle={International Conference on Learning Representations},
  year={2023}
}

@article{wang2024subjective,
  title={On Subjective Uncertainty Quantification and Calibration in Natural Language Generation},
  author={Wang, Ziyu and Holmes, Chris},
  journal={arXiv:2406.05213},
  year={2024}
}

@article{yang2024verbalized,
  title={On Verbalized Confidence Scores for {LLM}s},
  author={Yang, Daniel and Tsai, Yao-Hung Hubert and Yamada, Makoto},
  journal={arXiv:2412.14737},
  year={2024}
}

@inproceedings{lin2024generating,
  title={Generating with confidence: Uncertainty quantification for black-box large language models},
  author={Lin, Zhen and Trivedi, Shubhendu and Sun, Jimeng},
  booktitle={Transactions on Machine Learning Research},
  year={2024}
}

@article{grewal2024improving,
  title={Improving Uncertainty Quantification in Large Language Models via Semantic Embeddings},
  author={Grewal, Yashvir S and Bonilla, Edwin V and Bui, Thang D},
  journal={arXiv:2410.22685},
  year={2024}
}

@inproceedings{hou2024decomposing,
  title={Decomposing uncertainty for large language models through input clarification ensembling},
  author={Hou, Bairu and Liu, Yujian and Qian, Kaizhi and Andreas, Jacob and Chang, Shiyu and Zhang, Yang},
  booktitle={International Conference on Machine Learning},
  year={2024}
}

@inproceedings{gao2024spuq,
    title = "{SPUQ}: Perturbation-Based Uncertainty Quantification for Large Language Models",
    author = "Gao, Xiang  and
      Zhang, Jiaxin  and
      Mouatadid, Lalla  and
      Das, Kamalika",
    booktitle = "Proceedings of the 18th Conference of the European Chapter of the Association for Computational Linguistics (Volume 1: Long Papers)",
    year = "2024",
    doi = "10.18653/v1/2024.eacl-long.143",
}

@inproceedings{qiu2024semantic,
  title={Semantic density: Uncertainty quantification for large language models through confidence measurement in semantic space},
  author={Qiu, Xin and Miikkulainen, Risto},
  booktitle={Advances in Neural Information Processing Systems},
  year={2024}
}

@inproceedings{nikitin2024kernel,
  title={Kernel language entropy: Fine-grained uncertainty quantification for {LLM}s from semantic similarities},
  author={Nikitin, Alexander and Kossen, Jannik and Gal, Yarin and Marttinen, Pekka},
  booktitle={Advances in Neural Information Processing Systems},
  year={2024}
}

@article{ling2024uncertainty,
  title={Uncertainty quantification for in-context learning of large language models},
  author={Ling, Chen and Zhao, Xujiang and Zhang, Xuchao and Cheng, Wei and Liu, Yanchi and Sun, Yiyou and Oishi, Mika and Osaki, Takao and Matsuda, Katsushi and Ji, Jie and others},
  journal={arXiv:2402.10189},
  year={2024}
}

@inproceedings{jesson2024estimating,
  title={Estimating the hallucination rate of generative {AI}},
  author={Jesson, Andrew and Beltran Velez, Nicolas and Chu, Quentin and Karlekar, Sweta and Kossen, Jannik and Gal, Yarin and Cunningham, John P and Blei, David},
  booktitle={Advances in Neural Information Processing Systems},
  year={2024}
}

@inproceedings{lakshminarayanan2017simple,
  title={Simple and scalable predictive uncertainty estimation using deep ensembles},
  author={Lakshminarayanan, Balaji and Pritzel, Alexander and Blundell, Charles},
  booktitle={Advances in Neural Information Processing Systems},
  year={2017}
}

@inproceedings{
ritter2018,
title={{A Scalable Laplace Approximation for Neural Networks}},
author={Hippolyt Ritter and Aleksandar Botev and David Barber},
booktitle={International Conference on Learning Representations},
year={2018},
}

@inproceedings{kristiadi2020being,
  title={Being {B}ayesian, even just a bit, fixes overconfidence in relu networks},
  author={Kristiadi, Agustinus and Hein, Matthias and Hennig, Philipp},
  booktitle = {International Conference on Machine Learning},
  year={2020}
}

@inproceedings{daxberger2021laplace,
  title={Laplace redux-effortless {B}ayesian deep learning},
  author={Daxberger, Erik and Kristiadi, Agustinus and Immer, Alexander and Eschenhagen, Runa and Bauer, Matthias and Hennig, Philipp},
  booktitle={Advances in Neural Information Processing Systems},
  year={2021}
}

@article{wang2023lora,
  title={LoRA ensembles for large language model fine-tuning},
  author={Wang, Xi and Aitchison, Laurence and Rudolph, Maja},
  journal={arXiv:2310.00035},
  year={2023}
}

@inproceedings{yang2024bayesian,
  title={Bayesian low-rank adaptation for large language models},
  author={Yang, Adam X and Robeyns, Maxime and Wang, Xi and Aitchison, Laurence},
  booktitle={International Conference on Learning Representations},
  year={2024}
}

@inproceedings{faria2024quest,
  title={{QUEST}: Quality-Aware Metropolis-Hastings Sampling for Machine Translation},
  author={Faria, Gon{\c{c}}alo and Agrawal, Sweta and Farinhas, Ant{\'o}nio and Rei, Ricardo and de Souza, Jos{\'e} and Martins, Andr{\'e}},
  booktitle={Advances in Neural Information Processing Systems},
  year={2024}
}

@inproceedings{zhao2024knowing,
  title={Knowing what llms do not know: A simple yet effective self-detection method},
  author={Zhao, Yukun and Yan, Lingyong and Sun, Weiwei and Xing, Guoliang and Meng, Chong and Wang, Shuaiqiang and Cheng, Zhicong and Ren, Zhaochun and Yin, Dawei},
  booktitle={Proceedings of the 2024 conference of the north American chapter of the Association for Computational Linguistics: Human language technologies (Volume 1: Long Papers)},
  pages={7051--7063},
  year={2024}
}

@inproceedings{feng2025rethinking,
  title={Rethinking LLM uncertainty: A multi-agent approach to estimating black-box model uncertainty},
  author={Feng, Yu and Htut, Phu Mon and Qi, Zheng and Xiao, Wei and Mager, Manuel and Pappas, Nikolaos and Halder, Kishaloy and Li, Yang and Benajiba, Yassine and Roth, Dan},
  booktitle={Findings of the Association for Computational Linguistics: EMNLP 2025},
  pages={12349--12375},
  year={2025}
}

@inproceedings{feng2025bird,
  title={BIRD: A Trustworthy Bayesian Inference Framework for Large Language Models},
  author={Feng, Yu and Zhou, Ben and Lin, Weidong and Roth, Dan},
  booktitle={The Thirteenth International Conference on Learning Representations},
  year={2025}
}

@article{abbasi2024believe,
  title={To believe or not to believe your llm: Iterative prompting for estimating epistemic uncertainty},
  author={Abbasi Yadkori, Yasin and Kuzborskij, Ilja and Gy{\"o}rgy, Andr{\'a}s and Szepesvari, Csaba},
  journal={Advances in Neural Information Processing Systems},
  volume={37},
  pages={58077--58117},
  year={2024}
}

@inproceedings{tonolini2024bayesian,
  title={Bayesian prompt ensembles: Model uncertainty estimation for black-box large language models},
  author={Tonolini, Francesco and Aletras, Nikolaos and Massiah, Jordan and Kazai, Gabriella},
  booktitle={Findings of the Association for Computational Linguistics ACL 2024},
  pages={12229--12272},
  year={2024}
}

@article{zhang2024understanding,
  title={Understanding the relationship between prompts and response uncertainty in large language models},
  author={Zhang, Ze Yu and Verma, Arun and Doshi-Velez, Finale and Low, Bryan Kian Hsiang},
  journal={arXiv preprint arXiv:2407.14845},
  year={2024}
}
